\documentclass{article} % For LaTeX2e
\usepackage{iclr2026_conference,times}

\usepackage{amsmath,amsfonts,bm}

\def\eqref#1{equation~\ref{#1}}
\def\1{\bm{1}}

\DeclareMathAlphabet{\mathsfit}{\encodingdefault}{\sfdefault}{m}{sl}
\SetMathAlphabet{\mathsfit}{bold}{\encodingdefault}{\sfdefault}{bx}{n}

\usepackage{hyperref}
\usepackage{url}
\usepackage{xcolor}%
\usepackage{textcomp}%
\usepackage{manyfoot}%
\usepackage{booktabs}%
\usepackage{listings}%
\usepackage{mathtools}
\usepackage{nicefrac}       % compact symbols for 1/2, etc.
\usepackage{microtype}      % microtypography
\usepackage{xcolor}         % colors
\definecolor{rblue}{rgb}{0,0.5,1}
\usepackage{xcolor}
\usepackage{framed}
\usepackage{algorithm}
\usepackage{algorithmicx}
\usepackage[rightComments=true,italicComments=true,beginComment=//~]{algpseudocodex}
\usepackage{wrapfig,lipsum,booktabs}

\definecolor{darkgray}{gray}{0.3}

\newcommand{\sComment}[1]{\Comment{\small #1}}

\usepackage{amsmath}
\usepackage{booktabs}
\usepackage{multirow}
\usepackage{colortbl}
\usepackage{booktabs}
\usepackage{hhline}
\usepackage{subcaption}
\usepackage{tikz}
\usepackage{lipsum}
\usepackage{paralist}
\usepackage{color}

\makeatletter
\@ifundefined{isChecklistMainFile}{
  \newif\ifreproStandalone
  \reproStandalonetrue
}{
  \newif\ifreproStandalone
  \reproStandalonefalse
}
\makeatother

\makeatletter
\newcommand{\ie}{\emph{i.e.}\@ifnextchar.{\!\@gobble}{}}
\newcommand{\eg}{\emph{e.g.}\@ifnextchar.{\!\@gobble}{}}
\newcommand{\etc}{etc\@ifnextchar.{}{.\@}}
\makeatother

\usepackage{listings}

\definecolor{dkgreen}{rgb}{0,0.6,0}
\definecolor{gray}{rgb}{0.5,0.5,0.5}
\definecolor{mauve}{rgb}{0.58,0,0.82}
\definecolor{lightgray}{HTML}{DDDDDD}
\usepackage[rightComments=true,italicComments=true,beginComment=//~]{algpseudocodex}

\usepackage{newfloat}
\usepackage{listings}

\title{EReLiFM: Evidential Reliability-Aware Residual Flow Meta-Learning for Open-Set Domain Generalization under Noisy Labels}

\author{Kunyu Peng$^{1,2,}$\thanks{Correspondence: kunyu.peng@kit.edu, kailun.yang@hnu.edu.cn} \quad
Di Wen$^{1}$ \quad
Kailun Yang$^{3,*}$ \quad
Jia Fu$^{4,5}$ \quad
Yufan Chen$^{1}$\\
\textbf{Ruiping Liu}$^{1}$ \quad
\textbf{Jiamin Wu}$^{6,7}$ \quad
\textbf{Junwei Zheng}$^{1}$ \quad 
\textbf{M. Saquib Sarfraz}$^{1}$\\
\textbf{Luc Van Gool}$^{2}$ \quad
\textbf{Danda Pani Paudel}$^{2}$ \quad
\textbf{Rainer Stiefelhagen}$^{1}$\\
$^1$Karlsruhe Institute of Technology \quad
$^2$INSAIT, Sofia University ``St. Kliment Ohridski'' \\
$^3$Hunan University \quad
$^4$KTH Royal Institute of Technology \quad
$^5$RISE Research Institutes of Sweden\\
$^6$Shanghai Artificial Intelligence Laboratory \quad
$^7$The Chinese University of Hong Kong
}

\iclrfinalcopy % Uncomment for camera-ready version, but NOT for submission.
\begin{document}

\maketitle

\begin{abstract}
Open-Set Domain Generalization (OSDG) aims to enable deep learning models to recognize unseen categories in new domains, which is crucial for real-world applications. Label noise hinders open-set domain generalization by corrupting source-domain knowledge, making it harder to recognize known classes and reject unseen ones. While existing methods address OSDG under Noisy Labels (OSDG-NL) using hyperbolic prototype-guided meta-learning, they struggle to bridge domain gaps, especially with limited clean labeled data. In this paper, we propose Evidential Reliability-Aware Residual Flow Meta-Learning (EReLiFM). We first introduce an unsupervised two-stage evidential loss clustering method to promote label reliability awareness. Then, we propose a residual flow matching mechanism that models structured domain- and category-conditioned residuals, enabling diverse and uncertainty-aware transfer paths beyond interpolation-based augmentation. During this meta-learning process, the model is optimized such that the update direction on the clean set maximizes the loss decrease on the noisy set, using pseudo labels derived from the most confident predicted class for supervision. Experimental results show that EReLiFM outperforms existing methods on OSDG-NL, achieving state-of-the-art performance. The source code is available at \url{https://github.com/KPeng9510/ERELIFM}.
\end{abstract}

\section{Introduction}
Open-Set Domain Generalization (OSDG) tackles both domain and category shifts, requiring models to classify known categories while rejecting unseen ones. It is critical in dynamic applications such as healthcare~\cite{li2020domain}, security~\cite{busto2018open}, and autonomous driving~\cite{guo2022simt}, where new domains and categories often arise. Recent works employ meta-learning~\cite{wang2023generalizable, shu2021open} to simulate cross-domain tasks during training, improving adaptability to novel environments.
Yet, one can never expect the annotation to be $100\%$ correct. Label noise further complicates OSDG by compromising the reliability of knowledge learned from source domains. This challenges existing OSDG approaches as introduced in~\cite{peng2024mitigating}. Although label noise has been extensively studied in standard classification tasks, it remains largely unaddressed in OSDG. 

Existing techniques, such as relabeling~\cite{zhang2024badlabel,zheng2020error,li2024nicest}, data pruning~\cite{kim2021fine,karim2022unicon}, and loss-based noise-agnostic methods~\cite{xu2024skeleton,yue2024ctrl} focus on refining training data by correcting mislabeled instances or through selective optimization based on loss values. 
However, these methods do not address the additional challenge of adapting to unseen domains and distinguishing novel categories, which is essential in OSDG. 
\cite{peng2024mitigating} introduced novel benchmarks for the task of OSDG under Noisy Labels
(OSDG-NL) based on widely-used PACS~\cite{li2017deeper} and DigitsDG~\cite{zhou2020deep} datasets. Related approaches from both the OSDG and noisy label learning fields are evaluated as baselines.

HyProMeta~\cite{peng2024mitigating} serves as the first solution developed specifically targeting OSDG-NL, where hyperbolic prototypes are used to guide meta-learning optimization. Label noise agnostic meta-learning in HyProMeta is achieved by computing hyperbolic category prototypes to separate clean and noisy samples based on hyperbolic distances, correcting noisy labels using nearest prototypes, and augmenting training with a learnable prompt to enhance generalization to unseen categories.
However, prototype-based classification in HyProMeta is limited by sensitivity to noise and feature quality, which results in a negative effect on label noise diagnosis. Due to the limited number of clean samples and limited label-clean/noisy partition capability, HyProMeta suffers from unsatisfactory generalization performance, as less trustworthy a priori can be provided for the label-noise-agnostic meta-learning.

In this work, we propose a new method, \ie, Evidential Reliability-Aware Residual Flow Meta-Learning (EReLiFM). Our method introduces a new synergy between uncertainty-aware label reliability modeling and domain-category transfer modeling, which has not been explored in OSDG-NL.
Unlike prior works that either (i) separate clean/noisy samples using feature-space prototypes (HyProMeta) or (ii) rely on linear interpolation (MixUp) for augmentation, our method introduces a fundamentally different paradigm. First, we propose UTS-ELC, which leverages evidential loss trajectories to capture not only prediction errors but also their associated uncertainties, enabling more reliable clean/noisy separation across domains. Second, we introduce DC-CRFM, a flow-matching strategy conditioned on domain and category labels, which learns structured residuals rather than interpolations, thereby modeling diverse transfer paths between categories and domains. Finally, by integrating these two components within a meta-learning framework, we achieve principled decoupling of clean and noisy supervision, which is absent in prior methods. This combination enables EReLiFM to provide both uncertainty-aware noise diagnosis and diverse domain-category transfer modeling capabilities that neither clustering nor augmentation methods alone can offer.
Our approach achieves state-of-the-art results on the PACS~\cite{li2017deeper}, DigitsDG~\cite{zhou2020deep}, and TerraINC~\cite{beery2018recognition} datasets, showing its effectiveness in providing diverse cues to ensure correct optimization.

\section{Related Work}
\label{sec2}
\noindent\textbf{Noisy Label Learning.}
Accurate labels are crucial for deep learning models to acquire reliable information~\cite{xu2024skeleton}, while mislabeled data can mislead the optimization~\cite{cheng2020weakly}. To combat label noise, various strategies have been proposed: label corruption probabilities modeling~\cite{xia2019anchor, tanno2019learning, zhu2021clusterability, zhu2022beyond, li2022estimating}, re-weighting samples to adjust loss contributions~\cite{liu2015classification}, and detecting noisy labels before training~\cite{song2019selfie, wei2021smooth, chen2021beyond}.
TCL~\cite{huang2023twin} applies contrastive learning and Gaussian Mixture Models. 
Furthermore, noise-robust loss functions~\cite{liu2020peer, ma2020normalized, zhu2021second} and regularization tricks~\cite{wei2021open, cheng2021mitigating, liu2022robust} enhance model resilience. Methods like BadLabel~\cite{zhang2024badlabel} and LSL~\cite{kim2024learning} leverage label-flipping attacks and label structure, respectively. 
Notably, HyProMeta~\cite{peng2024mitigating} first introduces two benchmarks for the challenging OSDG-NL.

\noindent\textbf{Open-Set Domain Generalization.}
Open-Set Domain Generalization (OSDG) presents two interrelated challenges: domain generalization~\cite{wang2020heterogeneous, nam2021reducing, zhou2020domain, guo2023aloft, zhou2020learning, li2021progressive, li2021simple}, which trains models to transfer across source domains and the unseen, and open-set recognition~\cite{wang2024towards, zhao2023open, bao2021evidential,geng2020recent,peng2024navigating}, which aims to reject unknown categories with low confidence scores~\cite{fu2020learning, singha2024unknown, bose2023beyond, chen2021adversarial, li2018deep, zhao2022adaptive}. Although typically studied separately, OSDG explores strategies to address both challenges simultaneously. 
Previous work has investigated metric learning~\cite{katsumata2021open}, domain-augmented meta-learning~\cite{shu2021open}, and GAN-based data synthesis~\cite{bose2023beyond} to boost model robustness. 
Recently, formalized OSDG protocols~\cite{wang2023generalizable} have demonstrated the effectiveness of meta-learning in handling OSDG. 
HyProMeta~\cite{peng2024mitigating} focuses on hyperbolic prototypes to distinguish label-clean/noisy data, but is limited by the information scarcity of the limited label-clean samples. 
Flow-matching-based approaches~\cite{dao2023flow,gat2024discrete,klein2023equivariant,chen2023riemannian,eijkelboom2024variational} have gained attention for their effectiveness in optimal transportation between distributions and real-world applications. 
We propose EReLiFM, which integrates evidential-loss-based clean/noisy partitioning with domain- and category-conditioned residual flow in a meta-learning framework, achieving significant improvements over existing OSDG-NL methods.

\section{Methodology}
\subsection{Task Description}
\label{sec:task_description}
In this task, we consider a set containing $N_d$ domains $\mathcal{D} = \{d_1, d_2, ..., d_{N_d}\}$ and adopt the leave-one-out setting from~\cite{wang2023generalizable}, where a single domain $d_{t}$ is reserved for testing, while the remaining $\mathcal{D}_{S} = \mathcal{D}/\{d_{t}\}$ serve as source domains during training. 
The dataset's label set $\mathcal{Y}$ consists of $\mathcal{Y}_k$ (known categories in training) and $\mathcal{Y}_u$ (unseen categories in test), where $\mathcal{Y} = \mathcal{Y}_k \cup \mathcal{Y}_u$. For each pair of the sample $\mathbf{x}_{s}$ and label $\mathbf{y}_{s}$ in the source domain, $\mathbf{y}_s$ is converted to other known categories according to the different label noise settings to simulate the annotation error. Our aim is to achieve the best optimization when label noise exists in open-set domain generalization.

\subsection{EReLiFM}
\label{sec:EReLiFM}
In this work, we propose Evidential Reliability-Aware Residual Flow Meta-Learning (EReLiFM) to deal with noisy labels within the realm of OSDG, which will be elaborated in this subsection. OSDG leverages reliable cues from source domains and known categories to recognize unknown categories in unseen domains~\cite{peng2024advancing,wang2023generalizable}.
However, since label noise reduces the scale of reliable data, most of the existing works in the open-set domain generalization field deliver limited performance under label noise. 
Recognition of the data with label noise is critical to handle label noise for providing reliable optimization direction guidance for the deep learning model during optimization.
Existing work, \textit{i.e.}, HyProMeta~\cite{peng2024mitigating}, relies on clustering on embeddings for label noise agnostic learning, which is sensitive to outliers and feature quality, delivering limited performance. 
In this work, we optimize this process by proposing Unsupervised Two-Stage Evidential Loss Clustering (\textbf{UTS-ELC}), which separates label-clean/noisy data from a training dynamics perspective. We rely on evidential training dynamics instead of embeddings to achieve label noise diagnosis to avoid the sensitivity to outlier embeddings. 

Early works~\cite{han2018co,liu2021co} adopt multi-model joint optimization strategies based on the small-loss criterion. However, such approaches do not explicitly achieve a clear separation between clean and noisy labels. More recently, \cite{yue2024ctrl} introduce an unsupervised clustering strategy on recorded training dynamics to perform this separation. In contrast, our experiments show that a domain- and category-aware evidential loss leads to a more reliable distinction between clean and noisy sets under the open-set domain generalization scenario.

Despite this separation, training remains challenged by the scarcity of reliably annotated data. HyProMeta~\cite{peng2024mitigating} addresses this issue through cross-category MixUp and learnable prompts, thereby expanding the data scope to stabilize training. Yet, the diversity remains limited, since MixUp models only a single interpolation path between source and target data.

To mitigate domain shift, enhance the model’s sensitivity to diverse category transfers, and expand the scale of reliably annotated training data, we introduce \textbf{D}omain and \textbf{C}ategory \textbf{C}onditioned \textbf{R}esidual \textbf{F}low \textbf{M}atching (DC-CRFM). DC-CRFM generates diverse transfer paths by reconstructing domain- and category-residuals from random noise, conditioned on both domain and category labels. In this way, DC-CRFM explicitly models transitions across categories and domains, boosting generalization during training. Importantly, we train DC-CRFM on the clean subset identified by \textbf{UTS-ELC}.
Unlike MixUp, which interpolates between samples, DC-CRFM learns structured residuals across domains and categories. As demonstrated in our ablations (Tab.~\ref{tab:abl}), this design yields significant improvements over MixUp, evidencing that DC-CRFM is fundamentally distinct from interpolation-based augmentation.

Meta-learning has been proven effective for open-set domain generalization by constructing tailored meta-tasks to promote cross-domain generalization~\cite{wang2023generalizable,peng2024advancing,peng2024mitigating}. Building upon this insight, our main training framework adopts a meta-learning paradigm. Specifically, we define a new meta-training task over UTS-ELC-selected clean data and DC-CRFM-augmented clean data based on UTS-ELC selection. The optimized model from meta-training is then used to improve optimization on the noisy subset during meta-testing. Here, DC-CRFM plays a central role by enriching label-clean data with diverse category/domain transfer paths. Samples from the noisy set are supervised with high-confidence pseudo-labels via evidential learning, and regularized by cross-entropy loss against the original labels, thereby reinforcing consistency with the label-clean set.
Compared with HyProMeta~\cite{peng2024mitigating}, our meta-task differs in both meta-train and meta-test phases. In the meta-train stage, we exclusively rely on DC-CRFM augmented clean data, avoiding any optimization over noisy samples. In the meta-test stage, we focus solely on the noisy set: pseudo-labels are assigned via maximum-confidence predictions, and supervision is defined by a competition between pseudo-labels and original labels, as UTS-ELC does not guarantee perfect separation. To further account for uncertainty, evidential supervision is imposed on the pseudo-labels.

EReLiFM follows a simple pipeline of filter $\rightarrow$ expand $\rightarrow$ recycle: UTS-ELC filters clean data, DC-CRFM expands it with structured diversity, and evidential pseudo-labeling cautiously recycles noisy data, each step addressing a distinct bottleneck.
No prior work combines uncertainty-aware label noise diagnosis with domain-category flow residuals within a meta-learning framework.
We detail each component in the following sections.

\noindent\textbf{Unsupervised Two-Stage Evidential Loss Clustering.}
To mitigate the detrimental impact of label noise on the residual flow matching design, we first categorize the data based on their recorded evidential loss trajectories, as samples trained with incorrect labels typically exhibit higher loss, in accordance with Co-Teaching~\cite{han2018co}. While UTS-ELC builds on the intuition of loss trajectory clustering, our key novelty is the use of evidential uncertainty and domain/category-specific cues, which provide a more reasonable and reliable separation of clean/noisy data under OSDG. 

Initially, we train the backbone network on the entire dataset, despite the presence of label noise, while employing cyclic learning rates to improve convergence stability. 
Furthermore, we integrate evidential learning to enhance the model's generalization capability as Eq.~(\ref{eq:1}). Evidential learning enables models to estimate both predictions and their associated uncertainty, leading to more reliable and calibrated predictions.
\begin{equation}
\label{eq:1}
    \mathcal{L}_{EL} = \sum_{i=1}^\mathcal{C} \left[\mathbf{y}_{i} \left(\log S_{EL} - \log( \mathbf{M}_{\alpha}(\mathbf{x})_i + 1) \right)\right],
\end{equation}
where $S_{EL}=\sum_{i=1}^\mathcal{C} (\text{Dir}(p_{pred}|\mathbf{M}_{\alpha}(\mathbf{x})_i + 1))$ denotes the strength of a Dirichlet distribution, $\mathbf{M}_{\alpha}$ indicates the backbone, $\mathbf{y}_i$ is the one-hot annotation of sample $\mathbf{x}$ from class $i$, $p_{pred}$ is the predicted probability, and $\mathcal{C}$ is the class number.

During training, the evidential learning loss is recorded for each sample at every epoch. For a given sample $\mathbf{x}$, the recorded loss is represented as $\mathbf{l} = \left[l_1, l_2, ..., l_{N_e}\right]$, where $N_e$ denotes the total number of epochs.
We then construct a new feature set based on the recorded losses for each sample, formulated as $\mathcal{X}=\{\mathbf{l}_i| \mathbf{x}_i \in \mathcal{T}\}$, where $\mathcal{T}$ represents the entire training set.
To differentiate samples with and without label noise, we apply the unsupervised clustering method, FINCH~\cite{finch}, to the loss feature set, facilitating an initial hierarchical clustering process, according to Eq.~(\ref{eq:2}) and Eq.~(\ref{eq:3}).
\begin{equation}
\label{eq:2}
    \Omega_d^y = \{\omega_1, \omega_2, ..., \omega_{N_p}\} \leftarrow \text{FINCH}(\mathcal{X}_d^y),
\end{equation}
\begin{equation}
\label{eq:3}
    \Omega =\{\Omega_d^y |  d \in \mathcal{D}_S, y \in \mathcal{Y}_k\}, 
\end{equation}
where $\Omega$ represents the complete set of partitions, and $\mathcal{X}_d^y$ and $\Omega_d^y$ denote the loss set and the set of unsupervised cluster partitions for domain $d$ and class $y$, respectively.

Next, we construct a new set by computing the average of the samples within each domain and category for each partition according to Eq.~\ref{eq:4}. 
\begin{equation}
\label{eq:4}
\hat{\mathcal{X}}_{d}^{y} = \{\mu(\omega_1(\mathcal{X}_{d}^{y})), \mu(\omega_2(\mathcal{X}_{d}^{y})),..., \mu(\omega_{N_p}(\mathcal{X}_{d}^{y}))\},
\end{equation}
where $\mu(\cdot)$ denotes the averaging operation, $N_p$ denotes the total number of partitions clustered by the first level results on FINCH~\cite{finch}, and $\hat{\mathcal{X}}_{d}^{y}$ represents the resultant score set.

Finally, a Gaussian Mixture Model (GMM) based classifier (with two Gaussian components) is applied to perform a binary classification on the score set. This facilitates a threshold-free partitioning of the training data. The GMM class with the lower average loss is identified as the label-clean set (\ie, $ \hat{\mathcal{X}}_{d}^{(y,c)}$) and the other is denoted as noisy set (\ie, $\hat{\mathcal{X}}_{d}^{(y,n)}$), as Eq.~(\ref{eq:5}).
\begin{equation}
\label{eq:5}
    \hat{\mathcal{X}}_{d}^{(y,c)}, \hat{\mathcal{X}}_{d}^{(y,n)} =  \text{GMM}(\hat{\mathcal{X}}_{d}),s.t. ~\mu(\hat{\mathcal{X}}_{d}^{(y,c)}) < \mu(\hat{\mathcal{X}}_{d}^{(y,n)}).
\end{equation}
We then obtain the corresponding dataset according to the aforementioned partition manner, where we use $\mathcal{T}_{clean}$ and $\mathcal{T}_{noisy}$ to denote the clean set and noisy set, respectively.

\noindent\textbf{Domain and Category Conditioned Residual Flow Matching.}
Flow matching is a technique in machine learning that aligns feature distributions between source and target domains~\cite{lipman2022flow}. It appears as an efficient alternative compared with diffusion models~\cite{Ho2020DenoisingDP} for data generation, where methods leveraging straight flows are introduced by~\cite{liu2022flow}. 
\begin{algorithm}[t]
    \caption{Training with EReLiFM.}
    \label{algorithm}
    \renewcommand{\thealgorithm}{}
    \begin{algorithmic}[1]

\Require
$\mathcal{D}_S$: source domain set; 
    $\mathcal{Y}_k$: known category set;
    $\mathbf{M}_{\alpha}$: neural network backbone;
    $\mathbf{M}_{\gamma}$: flow matching model;
    $\mathcal{L}_{CE}$: cross entropy loss;
    $\mathcal{L}_{EL}$: evidential learning loss;
    $\mathcal{T}$: dataset with label noise; $\mathbf{r}_0$: random Gaussian noise.
\State Dataset separation, $\mathcal{T}_{clean}, \mathcal{T}_{noisy} \leftarrow \textbf{UTS-ELC}(\mathcal{T})$.
\State Train $\mathbf{M}_{\gamma}$ on $\mathcal{T}_{clean}$ using domain and category residuals, conditioned by classes and domains.
\While{not converged}
\LComment{Meta-Training Stage}
\State  $\mathbf{B}_{clean} \leftarrow Iter(\mathcal{T}_{clean})$, with domain label and category label $\mathbf{y}_{d}$ and $\mathbf{y}_{c}$.
\State Sample $\hat{\mathbf{y}}_c \leftarrow \mathcal{Y}/\{\mathbf{y}_c\}$, and $\hat{\mathbf{y}}_d \leftarrow \mathcal{D}_S/\{\mathbf{y}_d\}$.
\State $\mathbf{R}_{d} \leftarrow \mathbf{M}_{\gamma}(\mathbf{r}_0, (\mathbf{y}_c, \mathbf{y}_c), (\mathbf{y}_d, \hat{\mathbf{y}}_d))$, generate domain residual.
\State $\mathbf{B}_{dr} \leftarrow Add(\mathbf{B}_{clean}, \mathbf{R}_{d})$, merge domain residual.
\State Assign $\mathbf{y}_{c} \rightarrow \mathbf{y}_{dr}$ for $\mathbf{B}_{dr}$.
\State $\mathbf{R}_{c} \leftarrow \mathbf{M}_{\gamma}(\mathbf{n}_0, (\mathbf{y}_c, \hat{\mathbf{y}}_c), (\mathbf{y}_d, \mathbf{y}_d))$, generate category residual.
\State $\mathbf{B}_{cr} \leftarrow Add(\mathbf{B}_{clean}, \mathbf{R}_{c})$, merge categorical residual.
\State Assign $\mathbf{y}_{a} \rightarrow \mathbf{y}_{cr}$ for $\mathbf{B}_{cr}$, where $\mathbf{y}_a$ denotes an additional class beyond known classes.
\State Update parameters based on $\mathcal{L}_{m-train} = \mathcal{L}_{CE}(\mathbf{B}_{clean}, \mathbf{y}_c) + \mathcal{L}_{CE}(\mathbf{B}_{dr}, \mathbf{y}_{dr}) +  \mathcal{L}_{CE}(\mathbf{B}_{cr}, \mathbf{y}_{cr})$.
\LComment{Meta-Test Stage}
\State  $\mathbf{B}_{noisy} \leftarrow Iter(\mathcal{T}_{noisy})$ with category label $\mathbf{y}_{nc}$.
\State $\mathbf{y}_{pseudo} = ArgMax(\mathbf{M}_{\alpha}(\mathbf{B}_{noisy}))$.
\State $\mathcal{L}_{m-test} = \mathcal{L}_{EL}(\mathbf{B}_{noisy}, \mathbf{y}_{pseudo}) + \mathcal{L}_{CE}(\mathbf{B}_{noisy}, \mathbf{y}_{nc})$ .
\State $\text{UpdateParameters}(\mathcal{L}_{m-test} + \mathcal{L}_{m-train})$. \sComment{Final Parameter Update}
\EndWhile
\end{algorithmic}
\end{algorithm}

In our work, we propose a domain and category conditioned residual flow matching strategy to enrich the paths across different domains and categories based on a clean label set $\mathcal{T}_{clean}$. 
Domain residuals represent the visual differences between samples of the same category from different domains, while category residuals capture discrepancies between different categories within the same domain. We use our proposed conditioned flow matching to generate category and domain residuals. 

To enrich cross-domain and cross-category transfer, we propose \textbf{D}omain and \textbf{C}ategory \textbf{C}onditioned \textbf{R}esidual \textbf{F}low \textbf{M}atching (DC-CRFM), a conditioned variant of flow matching that learns residual distributions rather than directly generating samples. Given a source sample $\mathbf{I}_{\text{s}}$, a target sample $\mathbf{I}_{\text{t}}$ and a condition $\mathbf{q}$ (\eg, source$\to$target domain/category pair), RFM draws a residual $\mathbf{r}_1 = \mathbf{I}_{\text{t}}-\mathbf{I}_{\text{s}}  \sim p_r^{(\mathbf{q})}$ via a probability-flow ODE driven by a conditioned vector field $f_\theta$. Training is depicted in Eq.~\ref{eq:fm}.
\begin{equation}
\label{eq:fm}
\mathcal{L}_{\mathrm{RFM}} = \mathbb{E}_{(\mathbf{q},\,\mathbf{r}_0,\,\mathbf{r}_1,\,t)}\big[\| f_\theta(\mathbf{r}_t, t, \psi(\mathbf{q})) - (\mathbf{r}_1 - \mathbf{r}_0)\|_2^2\big], 
\quad \mathbf{r}_t = (1-t)\mathbf{r}_0 + t \mathbf{r}_1,
\end{equation} 
where $\mathbf{r}_0 \sim \mathcal{N}(0,1)$ (Gaussian distribution), $\mathbf{r}_1 \sim p_r^{(\mathbf{q})}$, $t \sim \mathcal{U}(0,1)$ (normal distribution), and $\psi(\mathbf{q})$ encodes the condition. At inference, integrating $\tfrac{d\mathbf{r}}{dt}=f_\theta(\mathbf{r},t,\psi(\mathbf{q}))$ from noise $\mathbf{r}_0\!\sim\!\mathcal{N}(0,1)$ yields $\mathbf{r}\!\sim\!p_r^{(\mathbf{q})}$, which is then added to $\mathbf{I}_{\text{s}}$ to form an augmented sample $\mathbf{I}_{\text{aug}} = \mathbf{I}_{\text{s}} + \mathbf{r}$. This design captures structured residual transitions between domains and categories, in contrast to simple interpolations such as MixUp~\cite{zhou2020domain,peng2024mitigating}. 

\noindent\textbf{Evidential Reliability-Aware Residual Flow Meta-Learning.} 
Through this process, we obtain a flow matching model that generates domain and category residuals while distinguishing clean from noisy data. These components are integrated into meta-learning for denoising and improving generalization in OSDG, as outlined in Alg.~\ref{algorithm}.

We first separate clean and noisy data using UTS-ELC, then train $\mathbf{M}_{\alpha}$ on $\mathcal{T}_{clean}$ with residual augmentation. In the meta-train stage, for each $\mathbf{B}_{clean}$ with labels $(\mathbf{y}_c,\mathbf{y}_d)$, we sample $(\hat{\mathbf{y}}_c,\hat{\mathbf{y}}_d)$ to generate residuals, producing $\mathbf{B}_{dr}$ (domain residual, supervised by $(\mathbf{y}_c,\mathbf{y}_d)$) and $\mathbf{B}_{cr}$ (category residual, supervised by an additional class $\mathbf{y}_a$). We assign $\mathbf{y}_{a} \rightarrow \mathbf{y}_{cr}$ for $\mathbf{B}_{cr}$. The model is updated with $\mathcal{L}_{m\text{-}train} = \mathcal{L}_{CE}(\mathbf{B}_{clean},\mathbf{y}_c) + \mathcal{L}_{CE}(\mathbf{B}_{dr},\mathbf{y}_{dr}) + \mathcal{L}_{CE}(\mathbf{B}_{cr},\mathbf{y}_{cr})$.

In the meta-test stage, noisy samples $\mathbf{B}_{noisy}$ are optimized via competition between the original label $\mathbf{y}_{nc}$ and a pseudo-label $\mathbf{y}_{pseudo} = ArgMax(M_{\alpha}(\mathbf{B}_{noisy}))$, with evidential regularization. An auxiliary cross-entropy term ensures that useful cues can still be extracted from misclassified clean samples. The final loss combines both stages, $\mathcal{L}_{m\text{-}train}+\mathcal{L}_{m\text{-}test}$, ensuring robust optimization with reliable supervision. This pipeline strengthens cross-domain generalization while also improving recognition of out-of-distribution categories.
Overall, clean/noisy separation via evidential training dynamics enables reliable residual flow training, while flow-augmented clean data and noisy samples are optimized separately in meta-train and meta-test to ensure robust learning.

\section{Experiments}
\begin{table*}[t!]

\centering
\resizebox{1.\linewidth}{!}{
\begin{tabular}{l|ccc|ccc|ccc|ccc|ccc}
\toprule
& \multicolumn{3}{c|}{\textbf{Photo (P)}} & \multicolumn{3}{c|}{\textbf{Art (A)}} & \multicolumn{3}{c|}{\textbf{Cartoon (C)}} & \multicolumn{3}{c|}{\textbf{Sketch (S)}} & \multicolumn{3}{c}{\textbf{Avg}} \\
\textbf{Method} & Acc & H-score & \cellcolor{gray!25}OSCR & Acc & H-score & \cellcolor{gray!25}OSCR & Acc & H-score & \cellcolor{gray!25}OSCR & Acc & H-score & \cellcolor{gray!25}OSCR & Acc & H-score & \cellcolor{gray!25}OSCR \\

\midrule

TCL~\cite{huang2023twin}  & 58.32 &	59.21&\cellcolor{gray!25}51.72  & 53.66 & 48.28 & \cellcolor{gray!25}42.91 & 46.78 & 38.29 & \cellcolor{gray!25}31.74 & 31.55 & 22.88 & \cellcolor{gray!25}24.30 & 47.58 & 42.17 & \cellcolor{gray!25}37.67 \\

NPN~\cite{sheng2024adaptive}  & 64.30 & 70.87 & \cellcolor{gray!25}61.99 & 51.66 & 52.10 &  \cellcolor{gray!25}45.40 & 55.65 & 44.88 & \cellcolor{gray!25}38.64 & 35.58 & 22.35 & \cellcolor{gray!25}25.86 & 51.80 & 47.55 & \cellcolor{gray!25}42.97 \\

BadLabel~\cite{zhang2024badlabel}  &54.93  &55.73  & \cellcolor{gray!25}48.24 & 53.72 & 53.25 & \cellcolor{gray!25}46.55 & 50.23 & 55.36 & \cellcolor{gray!25}45.70 & 31.55 & 21.84 & \cellcolor{gray!25}28.38 & 47.61 & 46.55 & \cellcolor{gray!25}42.22 \\

DISC~\cite{li2023disc} &53.47  &56.13  & \cellcolor{gray!25}47.22 & 54.47 & 47.46 & \cellcolor{gray!25}43.48 & 53.27 & 53.97 & \cellcolor{gray!25}44.33 & 24.01 & 16.75 & \cellcolor{gray!25}11.52 & 46.31 & 43.58 & \cellcolor{gray!25}36.64 \\
LSL~\cite{kim2024learning}  & 58.97 & 58.93 & \cellcolor{gray!25}52.15 & 49.97 & 48.17 & \cellcolor{gray!25}39.20 & 47.50 & 44.07 & \cellcolor{gray!25}34.63 & 30.59 & 12.44 & \cellcolor{gray!25}16.81 & 46.76 & 40.90 & \cellcolor{gray!25}35.70 \\
PLM~\cite{zhao2024estimating}  & 55.57 &42.33  & \cellcolor{gray!25}38.27 & 41.78 & 43.09 & \cellcolor{gray!25}32.95 & 45.75 & 40.44 & \cellcolor{gray!25}33.26 & 33.27 & 12.11 & \cellcolor{gray!25}15.46 & 38.33 & 34.49 & \cellcolor{gray!25}29.99 \\
\bottomrule
\end{tabular}

}
\vskip-2ex

\caption{Results (\%) of PACS on ResNet18. The open-set ratio is $6{:}1$ and symmetric label noise is with ratio $20\%$.}
\label{tab:pacs_res18_noise_20}
\vskip-3.5ex
\end{table*}

\begin{table*}[t!]

\centering
\resizebox{1.\linewidth}{!}{
\begin{tabular}{l|ccc|ccc|ccc|ccc|ccc}
\toprule
& \multicolumn{3}{c|}{\textbf{Photo (P)}} & \multicolumn{3}{c|}{\textbf{Art (A)}} & \multicolumn{3}{c|}{\textbf{Cartoon (C)}} & \multicolumn{3}{c|}{\textbf{Sketch (S)}} & \multicolumn{3}{c}{\textbf{Avg}} \\
\textbf{Method} & Acc & H-score & \cellcolor{gray!25}OSCR & Acc & H-score & \cellcolor{gray!25}OSCR & Acc & H-score & \cellcolor{gray!25}OSCR & Acc & H-score & \cellcolor{gray!25}OSCR & Acc & H-score & \cellcolor{gray!25}OSCR \\

\midrule

TCL~\cite{huang2023twin}  & 54.68 & 52.40 & \cellcolor{gray!25}46.51 & 52.78 & 22.65 & \cellcolor{gray!25}30.94 & 47.19 & 37.73 & \cellcolor{gray!25}34.89 & 26.33 & 9.83 & \cellcolor{gray!25}7.62 & 45.25 & 30.65 & \cellcolor{gray!25}29.99 \\

NPN~\cite{sheng2024adaptive} & 48.38 & 38.12 & \cellcolor{gray!25}33.55 & 35.71 & 32.33 & \cellcolor{gray!25}24.47 & 38.94 & 26.88 & \cellcolor{gray!25}18.60 & 26.93 & 27.59 & \cellcolor{gray!25}18.96 & 37.49 & 31.23 & \cellcolor{gray!25}23.90 \\

BadLabel~\cite{zhang2024badlabel} & 46.20 & 57.45 & \cellcolor{gray!25}45.07 & 45.34 & 47.29 & \cellcolor{gray!25}37.89 & 35.17 & 43.35 & \cellcolor{gray!25}32.14 & 28.40 & 26.95 & \cellcolor{gray!25}15.71 & 38.78 & 43.76 & \cellcolor{gray!25}32.70 \\

DISC~\cite{li2023disc} & 52.52 & 56.07 & \cellcolor{gray!25}50.55 & 46.84 & 31.91 & \cellcolor{gray!25}30.35 & 28.47 & 28.28 & \cellcolor{gray!25}19.97 & 30.83 & 25.63 & \cellcolor{gray!25}24.78 & 39.67 & 35.47 & \cellcolor{gray!25}31.41 \\

LSL~\cite{kim2024learning}  & 41.36 & 30.83 & \cellcolor{gray!25}20.27 & 42.28 & 39.78 & \cellcolor{gray!25}31.40 & 42.39 & 37.59 & \cellcolor{gray!25}30.89 & 26.90 & 15.40 & \cellcolor{gray!25}7.42 & 38.23 & 30.90 & \cellcolor{gray!25}22.50 \\
PLM~\cite{zhao2024estimating}  &55.57  &42.33  & \cellcolor{gray!25}38.27 & 39.21 & 27.81 & \cellcolor{gray!25}24.93 & 33.01 & 27.81 & \cellcolor{gray!25}21.49 & 25.52 & 6.65 & \cellcolor{gray!25}13.41 & 38.33 & 26.15 & \cellcolor{gray!25}24.53 \\
\bottomrule
\end{tabular}
}
\vskip-2ex
\caption{Results (\%) of PACS on ResNet18.
The open-set ratio is $6{:}1$ and symmetric label noise is with ratio $50\%$.}
\label{tab:pacs_res18_50}
\vskip-3.5ex
\end{table*}

\begin{table*}[t!]

\centering
\resizebox{1.\linewidth}{!}{
\begin{tabular}{l|ccc|ccc|ccc|ccc|ccc}
\toprule
& \multicolumn{3}{c|}{\textbf{Photo (P)}} & \multicolumn{3}{c|}{\textbf{Art (A)}} & \multicolumn{3}{c|}{\textbf{Cartoon (C)}} & \multicolumn{3}{c|}{\textbf{Sketch (S)}} & \multicolumn{3}{c}{\textbf{Avg}} \\
\textbf{Method} & Acc & H-score & \cellcolor{gray!25}OSCR & Acc & H-score & \cellcolor{gray!25}OSCR & Acc & H-score & \cellcolor{gray!25}OSCR & Acc & H-score & \cellcolor{gray!25}OSCR & Acc & H-score & \cellcolor{gray!25}OSCR \\

\midrule

TCL~\cite{huang2023twin}  &31.58&	25.81&\cellcolor{gray!25}17.39  & 27.08 & 26.60 & \cellcolor{gray!25}16.45 & 27.69 & 27.17 & \cellcolor{gray!25}17.66 & 21.20 & 8.52 & \cellcolor{gray!25}14.87 & 26.89 & 22.03 & \cellcolor{gray!25}16.59 \\

NPN~\cite{sheng2024adaptive} & 17.21 & 12.49 & \cellcolor{gray!25}10.18 & 24.27 & 12.67 & \cellcolor{gray!25}13.87 & 22.85 & 12.99 & \cellcolor{gray!25}10.85 & 19.66 & 4.31 & \cellcolor{gray!25}11.63 & 21.00 & 10.62 & \cellcolor{gray!25}11.63 \\

BadLabel~\cite{zhang2024badlabel}  & 22.62 & 14.11 & \cellcolor{gray!25}22.62 & 15.95 & 14.50 & \cellcolor{gray!25}10.16 & 19.39 & 24.34 & \cellcolor{gray!25}14.95 & 26.13 & 14.14 & \cellcolor{gray!25}17.96 & 21.02  & 16.77 & \cellcolor{gray!25}16.42 \\

DISC~\cite{li2023disc} & 22.05 & 19.53 & \cellcolor{gray!25}15.27 & 24.77 & 23.65 & \cellcolor{gray!25}17.53 & 27.13 & 22.49 & \cellcolor{gray!25}14.76 & 16.03 & 12.86 & \cellcolor{gray!25}10.00 & 22.05 & 19.63 & \cellcolor{gray!25}14.39 \\

LSL~\cite{kim2024learning}  & 18.58 & 22.82 & \cellcolor{gray!25}13.89 & 23.64 & 16.71 & \cellcolor{gray!25}14.56 & 15.37 & 15.84 & \cellcolor{gray!25}7.94 & 21.68 & 1.92 & \cellcolor{gray!25}8.26 & 19.82 & 14.32 & \cellcolor{gray!25}11.16 \\
PLM~\cite{zhao2024estimating}  &24.39  &8.71  & \cellcolor{gray!25}9.41 & 30.08 & 24.67 & \cellcolor{gray!25}17.85 &  20.94& 13.16 & \cellcolor{gray!25}12.21 & 21.76 & 21.26 & \cellcolor{gray!25}11.74 & 24.29 & 16.95 & \cellcolor{gray!25}12.80 \\
\bottomrule
\end{tabular}
}
\vskip-2ex
\caption{Results (\%) of PACS on ResNet18.
The open-set ratio is $6{:}1$ and symmetric label noise is with ratio $80\%$.}

\label{tab:pacs_res18_80}
\vskip-3.5ex
\end{table*}

\begin{table*}[t!]

\centering
\resizebox{1.\linewidth}{!}{
\begin{tabular}{l|ccc|ccc|ccc|ccc|ccc}
\toprule
& \multicolumn{3}{c|}{\textbf{Photo (P)}} & \multicolumn{3}{c|}{\textbf{Art (A)}} & \multicolumn{3}{c|}{\textbf{Cartoon (C)}} & \multicolumn{3}{c|}{\textbf{Sketch (S)}} & \multicolumn{3}{c}{\textbf{Avg}} \\
\textbf{Method} & Acc & H-score & \cellcolor{gray!25}OSCR & Acc & H-score & \cellcolor{gray!25}OSCR & Acc & H-score & \cellcolor{gray!25}OSCR & Acc & H-score & \cellcolor{gray!25}OSCR & Acc & H-score & \cellcolor{gray!25}OSCR \\

\midrule

TCL~\cite{huang2023twin}  & 15.83&3.33	&\cellcolor{gray!25}10.69    & 35.21 & 30.54 & \cellcolor{gray!25}20.96 & 26.46 & 24.24 & \cellcolor{gray!25}14.88 & 20.35 & 1.20 & \cellcolor{gray!25}2.63 & 24.46 & 14.83 & \cellcolor{gray!25}12.29 \\

NPN~\cite{sheng2024adaptive} & 44.43 & 37.82 & \cellcolor{gray!25}28.94 & 38.65 & 39.48 & \cellcolor{gray!25}23.11 & 31.92 & 20.12 & \cellcolor{gray!25}13.49 & 23.24 & 11.25 & \cellcolor{gray!25}4.29 & 34.56 & 27.17 & \cellcolor{gray!25}17.46 \\

BadLabel~\cite{zhang2024badlabel} & 37.16 & 44.56 & \cellcolor{gray!25}35.70 & 26.89 & 31.98 & \cellcolor{gray!25}24.77 & 31.10 & 34.52 & \cellcolor{gray!25}25.33 & 13.88 & 16.21 & \cellcolor{gray!25}20.38 & 27.26 & 31.82 & \cellcolor{gray!25}26.54 \\

DISC~\cite{li2023disc} & 46.20 & 42.31 & \cellcolor{gray!25}38.82 & 44.28 & 42.69 & \cellcolor{gray!25}34.41 & 42.70 & 35.22 & \cellcolor{gray!25}26.28 & 32.77 & 2.77 & \cellcolor{gray!25}22.37 & 41.49 & 30.75 & \cellcolor{gray!25}30.47 \\

LSL~\cite{kim2024learning}  & 27.30 & 19.63 & \cellcolor{gray!25}14.32 & 24.27 & 11.56 & \cellcolor{gray!25}13.18 & 25.63 & 14.77 & \cellcolor{gray!25}12.04 & 22.21 & 16.31 & \cellcolor{gray!25}25.07 & 24.85 & 15.57 & \cellcolor{gray!25}16.15 \\
PLM~\cite{zhao2024estimating}  &17.37  &23.20  & \cellcolor{gray!25}14.13 & 25.45 & 24.44 & \cellcolor{gray!25}15.04 & 20.06 & 9.68 & \cellcolor{gray!25}11.93 & 20.48 & 2.60 & \cellcolor{gray!25}3.24 & 20.84 & 14.98 & \cellcolor{gray!25}11.09 \\
\bottomrule
\end{tabular}

}
\vskip-2ex
\caption{Results (\%) of PACS on ResNet18.
The open-set ratio is $6{:}1$ and asymmetric label noise is with ratio $50\%$.}
\label{tab:pacs_res18_50_a}
\vskip-3ex
\end{table*}

\begin{table*}[t!]

\centering
\resizebox{1\linewidth}{!}{
\begin{tabular}{l|ccc|ccc|ccc|ccc}
\toprule
& \multicolumn{3}{c|}{\textbf{20\% sym}} & \multicolumn{3}{c|}{\textbf{50\% sym}} & \multicolumn{3}{c|}{\textbf{80 \% sym}} & \multicolumn{3}{c}{\textbf{50\% asym}}  \\
\textbf{Method} & Acc & H-score & \cellcolor{gray!25}OSCR & Acc & H-score & \cellcolor{gray!25}OSCR & Acc & H-score & \cellcolor{gray!25}OSCR & Acc & H-score & \cellcolor{gray!25}OSCR \\

\midrule

TCL~\cite{huang2023twin}& 52.47 & 55.28 & \cellcolor{gray!25} 46.85  & 50.19 & 43.92 & \cellcolor{gray!25} 42.02 & 23.31 & 18.44 & \cellcolor{gray!25} 12.26 & 38.91 & 39.94 & \cellcolor{gray!25}30.98 \\

NPN~\cite{sheng2024adaptive}&  47.68 & 44.04 & \cellcolor{gray!25} 38.51 & 32.06 & 29.07 & \cellcolor{gray!25} 23.83& 17.95 & 11.12 & \cellcolor{gray!25} 11.42  & 25.54 & 17.72 & \cellcolor{gray!25}13.20 \\

BadLabel~\cite{zhang2024badlabel}& 49.06 & 50.57 & \cellcolor{gray!25} 46.04  & 39.83 & 45.65 & \cellcolor{gray!25} 36.62& 20.92 & 22.11 & \cellcolor{gray!25} 19.63 & 32.27 & 41.03 & \cellcolor{gray!25}31.86\\

DISC~\cite{li2023disc} &  52.21 & 40.11 & \cellcolor{gray!25} 42.93 & 36.73 & 34.36 & \cellcolor{gray!25} 27.93& 22.77 & 9.89 & \cellcolor{gray!25} 12.67  & 28.99 & 13.55 & \cellcolor{gray!25}11.21 \\

LSL~\cite{kim2024learning}  & 52.96 & 48.54 & \cellcolor{gray!25} 49.12& 50.19 & 42.60 & \cellcolor{gray!25} 40.80  & 23.39 & 12.55 & \cellcolor{gray!25} 12.69& 35.84 & 23.88 & \cellcolor{gray!25}19.73\\
PLM~\cite{zhao2024estimating}  & 52.94 & 47.03 & \cellcolor{gray!25} 46.62 & 42.17 & 37.45 & \cellcolor{gray!25} 35.33& 26.40 & 15.88 & \cellcolor{gray!25} 9.72   & 24.78 & 23.62 & \cellcolor{gray!25}17.48\\
\bottomrule
\end{tabular}
}
\vskip-2ex
\caption{Results (\%) of PACS on ViT-Base.
The open-set ratio is $6{:}1$. The average domain performance is reported.}

\label{tab:pacs_vit_20}
\vskip-3ex
\end{table*}

\begin{table*}[t!]

\centering
\resizebox{1\linewidth}{!}{
\begin{tabular}{l|ccc|ccc|ccc|ccc}
\toprule
& \multicolumn{3}{c|}{\textbf{20\% sym}} & \multicolumn{3}{c|}{\textbf{50\% sym}} & \multicolumn{3}{c|}{\textbf{80\% sym}} & \multicolumn{3}{c}{\textbf{50\% asym}} \\
\textbf{Method} & Acc & H-score & \cellcolor{gray!25}OSCR & Acc & H-score & \cellcolor{gray!25}OSCR & Acc & H-score & \cellcolor{gray!25}OSCR & Acc & H-score & \cellcolor{gray!25}OSCR \\

\midrule
NPN~\cite{sheng2024adaptive} & 38.49 & 23.03 & \cellcolor{gray!25}29.43 & 42.66 & 21.91 & \cellcolor{gray!25}28.83& 17.43 & 8.02 & \cellcolor{gray!25}9.33 &58.39 & 29.98 & \cellcolor{gray!25}42.38 \\

BadLabel~\cite{zhang2024badlabel}  &  46.31 & 38.81 & \cellcolor{gray!25}44.12  & 36.70 & 26.94 & \cellcolor{gray!25}33.37& 17.44 & 5.40 & \cellcolor{gray!25}8.30& 38.05 & 33.01 & \cellcolor{gray!25}33.26\\

\midrule 

ODGNet~\cite{bose2023beyond}&  68.88 & 40.87 & \cellcolor{gray!25}51.07 & 60.90 & 30.62 & \cellcolor{gray!25}43.12  &17.17 & 11.31 & \cellcolor{gray!25} 9.52 &45.96&	26.46	  & \cellcolor{gray!25}32.41\\

MLDG \cite{chen2021adversarial} & 65.66 & 30.50 & \cellcolor{gray!25}49.93 & 44.95 & 33.85 & \cellcolor{gray!25}32.31 & 17.23 & 5.91 & \cellcolor{gray!25} 9.00&57.19 &	29.38	  & \cellcolor{gray!25}41.57\\

MEDIC-cls~\cite{wang2023generalizable} & 20.09 & 12.23 & \cellcolor{gray!25}8.94& 17.17 & 10.21 & \cellcolor{gray!25}5.51 & 18.21 & 7.34 & \cellcolor{gray!25}8.83 &16.44  &	11.10	  & \cellcolor{gray!25}7.37\\

MEDIC-bcls~\cite{wang2023generalizable} & 20.09 &  13.65 & \cellcolor{gray!25}6.76 & 17.17 & 12.80 & \cellcolor{gray!25}5.28 & 17.73 & 7.47 & \cellcolor{gray!25}8.30 &16.44  &	14.55	  & \cellcolor{gray!25}6.98 \\
\bottomrule
\end{tabular}}
\vskip-2ex
\caption{Results ($\%$) of DigitsDG on ConvNet. The open-set ratio is $6{:}4$. The average domain performance is reported.}
\label{tab:dg_20}
\vskip-3ex
%\vskip-2ex
\end{table*}

\subsection{Noisy Label Settings}
\label{sec:label_noise}
We adopt the setting of HyProMeta~\cite{peng2024mitigating} for OSDG-NL, incorporating \textbf{symmetric} and \textbf{asymmetric} label noise. \textbf{Symmetric noise} randomly reassigns class labels at predefined rates ($20\%$, $50\%$, $80\%$) without considering semantics. In contrast, \textbf{asymmetric noise} mislabels samples according to semantic similarity using BERT~\cite{BERT} for textual feature extraction and cosine similarity for class similarity computation. The asymmetric noise level is set to $50\%$.

\subsection{Datasets and Metrics}
We adopt OSDG protocols from MEDIC~\cite{wang2023generalizable} and HyProMeta~\cite{peng2024mitigating}, where training domains share the same categories. Evaluation is on three benchmarks: \textbf{PACS}~\cite{li2017deeper} ( \textit{photo}, \textit{art-painting}, \textit{cartoon}, \textit{sketch}), \textbf{DigitsDG}~\cite{zhou2020deep} (\textit{mnist}, \textit{mnist-m}, \textit{svhn}, \textit{syn}), and \textbf{TerraINC}~\cite{beery2018recognition} (reported in Tab.~\ref{tab:terrainc} in appendix). We follow the leave-one-domain-out setting~\cite{wang2023generalizable}, using OSCR as the primary metric, with H-score and Acc as secondary metrics.

\subsection{Implementation Details}
The experiments are all conducted by PyTorch2.0 on one NVIDIA A100 GPU. Training is limited to \(1 \times 10^4\) steps, utilizing the SGD optimizer with a learning rate (LR) of \(1 \times 10^{-3}\) and a batch size of $16$. A learning rate decay of \(1 \times 10^{-1}\) is applied after \(8 \times 10^3\) meta-training steps. During the residual flow matching training, DiT~\cite{Peebles2022DiT} is utilized as the backbone, where the training batch size is set as $128$. $N_{e}$ is chosen as $10$.
Regarding the feature learning backbones, the ConvNet~\cite{zhou2021domain} is employed as the backbone network on the DigitsDG dataset, following~\cite{zhou2021domain}. EReLiFM is only applied during training.
In inference, no DiT structure is required, and the prediction relies solely on the chosen backbone and a lightweight classification head. This ensures test-time efficiency. For reference, the backbones used have parameter counts of $\sim 11.7M$ (ResNet18), $\sim 86M$ (ViT-Base), and $\sim 1.4M$ (ConvNet). 

\subsection{Analysis of the Model Performance}
In Tab.~\ref{tab:pacs_res18_noise_20}, Tab.~\ref{tab:pacs_res18_50}, Tab.~\ref{tab:pacs_res18_80}, and Tab.~\ref{tab:pacs_res18_50_a}, we present performance comparisons between our proposed approach and other related methods. Among these, TCL~\cite{huang2023twin}, NPN~\cite{sheng2024adaptive}, BadLabel~\cite{zhang2024badlabel}, DISC~\cite{li2023disc}, LSL~\cite{kim2024learning}, and PLM~\cite{zhao2024estimating} focus on label noise learning, while MEDIC~\cite{wang2023generalizable}, MLDG~\cite{shu2019meta}, ARPL~\cite{chen2021adversarial}, MixStyle~\cite{zhou2020domain}, ODGNet~\cite{bose2023beyond}, SWAD~\cite{cha2021swad}, and EBiL-HaDS~\cite{peng2024advancing} specifically target open-set domain generalization. HyProMeta~\cite{peng2024mitigating} is the first work addressing the OSDG-NL problem, utilizing hyperbolic prototypes to guide meta-learning. Although HyProMeta achieves the best performance among existing baselines, its reliance on a limited number of label-clean samples from the source domains and known classes constrains the model’s generalization capability for OSDG-NL.

Compared to HyProMeta~\cite{peng2024mitigating}, our approach achieves $14.76\%$, $11.56\%$, $5.14\%$, and $11.99\%$ accuracy improvements, $5.63\%$, $4.68\%$, $9.14\%$, and $3.63\%$ H-score improvements, and $13.70\%$, $9.01\%$, $4.93\%$, and $8.20\%$ OSCR improvements on the PACS dataset~\cite{li2017deeper} using ResNet18~\cite{he2016deep} as the feature learning backbone, under symmetric label noise ratios of $20\%$, $50\%$, $80\%$, and asymmetric label noise ratio $50\%$, respectively.
These improvements stem from residual flow matching, which enriches cross-category/domain paths, and UTS-ELC, which reliably separates clean from noisy labels. This allows effective optimization on limited clean data, while evidential learning further extracts cues from noisy samples during meta-test. We also find larger gains on visually rich domains (\ie, \textit{photo}, \textit{art painting}, \textit{cartoon}) than on \textit{sketch}; under $80\%$ symmetric noise, OSCR improvement on \textit{sketch} is only $1.35\%$, indicating our method is most effective when visual features are preserved.

EReLiFM outperforms HyProMeta because it addresses the weaknesses of prototype-based alignment at multiple levels. First, evidential training dynamics clustering separates clean from noisy samples, ensuring that training is guided by reliability-aware representations rather than corrupted prototypes. Second, domain- and category-conditioned residual flow matching models the distributional transport across domains and categories, capturing richer variations than simple mean-level alignment. Finally, the proposed evidential reliability-aware residual flow meta-learning pipeline systematically leverages clean, augmented, and cautiously recycled noisy data to expand the range of training tasks, thereby narrowing the gap to unseen domains. Together, these components form a principled framework that is theoretically more robust than HyProMeta, which relies solely on prototype matching.

\subsection{Cross-Backbone Generalizability}
To assess the cross-backbone generalizability, we conduct experiments using the ViT-Base~\cite{dosovitskiy2021an} backbone on PACS~\cite{li2017deeper} under the four label noise settings, as presented in Tab.~\ref{tab:pacs_vit_20}.
We first observe that employing a larger transformer architecture leads to overall performance improvements across all methods. Notably, HyProMeta~\cite{peng2024mitigating} achieves $6.77\%$, $8.93\%$, $0.17\%$, and $5.64\%$ OSCR improvements when using ViT-Base compared to ResNet18~\cite{he2016deep}. Similar trends are observed in the performance of our proposed approach.
Compared to the current state-of-the-art method, \ie, HyProMeta~\cite{peng2024mitigating}, our approach achieves $13.42\%$, $13.12\%$, $4.57\%$, and $5.06\%$ accuracy improvements, $0.94\%$, $1.46\%$, $7.66\%$, and $7.14\%$ H-score improvements, and $6.27\%$, $10.30\%$, $9.70\%$, and $7.16\%$ OSCR improvements under symmetric label noise ratios of $20\%$, $50\%$, $80\%$, and asymmetric label noise ratio of $50\%$, respectively. Per-target domain results are reported in the appendix.

\subsection{Evaluation on Another Dataset}
We further evaluate the generalizability of our proposed approach on the DigitsDG dataset, with results presented in Tab.~\ref{tab:dg_20}. Several state-of-the-art methods with strong OSDG-NL performance are reported, including NPN~\cite{sheng2024adaptive}, BadLabel~\cite{zhang2024badlabel}, ODGNet~\cite{bose2023beyond}, MLDG~\cite{shu2019meta}, MEDIC~\cite{wang2023generalizable}, EBiL-HaDS~\cite{peng2024advancing}, and HyProMeta~\cite{peng2024mitigating}. Among these, HyProMeta achieves the highest OSCR, with $55.34\%$, $44.10\%$, $10.63\%$, and $48.79\%$ under symmetric label noise ratios of $20\%$, $50\%$, $80\%$, and asymmetric label noise ratio of $50\%$, respectively.
Our approach consistently outperforms HyProMeta~\cite{peng2024mitigating}, achieving $61.34\%$, $47.19\%$, $11.57\%$, and $50.97\%$ OSCR under the same noise settings. 
This improvement highlights the robustness of our method in handling noisy labels while ensuring effective generalization across domains.
Our approach benefits from residual flow matching, which enriches domain and category knowledge, and UTS-ELC, which improves clean-noisy label separation for robust meta-learning. Results confirm effectiveness across OSDG-NL datasets, including TerraInc (Tab.~\ref{tab:terrainc}), where our method outperforms HyProMeta. 
Unlike prototype-based~\cite{peng2024mitigating} or interpolation-based~\cite{zhou2020domain} methods, which assume clean feature geometry or linear transition paths, residual flows approximate probabilistic transport maps between distributions. This theoretically provides a richer and more faithful modeling of domain- and category-conditioned shifts, and enables a more generalizable model optimization, especially when combined with evidential uncertainty for reliability-aware supervision during the label-noise-aware meta-learning stage. Further details can be found in the appendix.

\subsection{Analysis of the Module Ablation}

\noindent\textbf{Ablation of the DC-CRFM.} 
The ablation results are shown in Tab.~\ref{tab:abl}. To evaluate the impact of DC-CRFM, we examine five model variants: \textit{w/o DC-CRFM}, \textit{w/o domain RA}, \textit{w/o category RA}, \textit{w/ mixup (replace DC-CRFM)}, and \textit{w/ DirectFM}. \textit{w/o DC-CRFM} removes residual flow matching from meta-learning, \textit{w/o domain RA} excludes augmentation of generated domain residuals, \textit{w/o category RA} omits category residual augmentation, and \textit{w/ mixup (replace DC-CRFM)} uses direct cross-domain and -class MixUp to replace DC-CRFM. Our results show that \textit{w/o DC-CRFM} leads to $10.97\%$ and $6.45\%$ OSCR drop on target domains \textit{mnist} and \textit{syn}, highlighting the significance of using our proposed category and domain-conditioned residual flow matching in meta-learning. 
\begin{wraptable}{t}{7cm}
\vskip-1.5ex
\resizebox{0.5\textwidth}{!}{\begin{tabular}{l|ccc|ccc}
\toprule
\multicolumn{1}{c}{\multirow{2}{*}{\textbf{Variants}}} & \multicolumn{3}{|c}{\textbf{mnist}}                                               & \multicolumn{3}{|c}{\textbf{syn}}                                                 \\
\multicolumn{1}{c}{}                                   & \multicolumn{1}{|l}{ACC} & \multicolumn{1}{l}{H-score} & \multicolumn{1}{l}{OSCR} & \multicolumn{1}{|l}{ACC} & \multicolumn{1}{l}{H-score} & \multicolumn{1}{l}{OSCR} \\
\midrule
w/o DC-CRFM                                              & 71.89                   & 17.80                        & \cellcolor{gray!25}58.91                    & 50.19                   & 39.04                       & \cellcolor{gray!25}33.19                    \\
w/o domain RA                                          & 76.03                   & 30.23                       & \cellcolor{gray!25}66.66                    & 50.17                   & 35.59                       & \cellcolor{gray!25}30.20                     \\
w/o category RA                                     & 73.17                   & 28.87                       & \cellcolor{gray!25}66.02                    & 53.42                   & 37.54                       & \cellcolor{gray!25}33.27                    \\
w/ mixup (replace DC-CRFM)                                              &       80.61  &   14.96                     & \cellcolor{gray!25}67.46                     &        37.44            & 2.14                      & \cellcolor{gray!25}22.06                   \\
w/ DirectFM & 75.44 & 62.33 & \cellcolor{gray!25}63.28 & 54.69& 34.56 & \cellcolor{gray!25}37.08\\
\midrule
w/o UTS-ELC in RFM  &  77.42 &  16.84     &   \cellcolor{gray!25}64.48  & 47.31   &  19.08     &  2\cellcolor{gray!25}9.26   \\
w/ UTS-LC in RFM                   & 78.92                   & 23.89                       & \cellcolor{gray!25}59.62                    & 39.19                   & 24.30                        & \cellcolor{gray!25}25.50                     \\
\midrule

w/o $\mathcal{L}_{EL}$ in meta-test                            & 78.42                   & 23.33                       & \cellcolor{gray!25}61.15                    & 52.58                   & 36.62                       & \cellcolor{gray!25}33.68                    \\
w/o $\mathcal{L}_{CE}$ in meta-test                             &           69.89         &           60.09          & \cellcolor{gray!25}56.24                    &      38.17              &              16.10          & \cellcolor{gray!25} 24.36                   \\
\midrule
\textbf{Ours}                                          & \textbf{85.97}          & \textbf{64.79}              & \cellcolor{gray!25}\textbf{69.88}           & \textbf{56.61}          & \textbf{41.60}               & \cellcolor{gray!25}\textbf{39.64}       \\ 
\bottomrule
\end{tabular}}
\vskip-2ex
\caption{Module ablation on the DigitsDG dataset, symmetric label noise with ratio $50\%$ is selected.}
\vskip-2ex
\label{tab:abl}
\end{wraptable} 
Additionally, our approach consistently outperforms \textit{w/o domain RA}, \textit{w/o category RA}, and \textit{w/ mixup (replace DC-CRFM)}, demonstrating the superior design of DC-CRFM for the OSDG-NL task. Notably, DC-CRFM consistently outperforms MixUp (\textit{w/ mixup (replace DC-CRFM)}) by large margins, confirming that flow matching is not a simple interpolation-based augmentation. Instead, it learns structured residuals conditioned on domains and categories, enabling richer and more transferable paths.

\noindent\textbf{Ablation of the clean/noisy dataset partition technique.}
We present two variants: \textit{w/o UTS-ELC in RFM} and \textit{w/ UTS-LC in RFM}. The former trains residual flow matching without label-clean/noisy data separation, while the latter excludes evidential learning loss. Our approach outperforms both variants by $>5\%$ OSCR, demonstrating the importance of proper label-clean/noisy data partitioning and the benefit of using evidential learning loss in the unsupervised loss clustering.

\noindent\textbf{Ablation of meta-learning task.} We further conduct another ablation regarding the meta-learning by removing the evidential pseudo-label supervision in the meta-test stage, indicated by \textit{w/o $\mathcal{L}_{EL}$}. Our proposed method contributes $8.73\%$ and $5.96\%$ performance gains in terms of OSCR, illustrating the superiority of using evidential pseudo-label supervision on the label-noisy set during the meta-training for the model optimization. On the other hand, the variant \textit{w/o $\mathcal{L}_{CE}$} shows a performance drop, indicating the importance of both losses. While $\mathcal{L}_{EL}$ enables label correction, $\mathcal{L}_{CE}$ helps extract useful cues from misassigned clean samples in the noisy set.

\section{Conclusion}
We present \textbf{EReLiFM}, a reliability-aware residual flow meta-learning framework for open-set domain generalization under noisy labels. By combining evidential clustering for clean/noisy data separation with domain- and category-conditioned flow matching, our method enhances data reliability and diversity for meta-learning. Experiments on multiple benchmarks confirm that EReLiFM achieves robust performance against label noise and strong generalization to unseen domains and categories.

\section*{Reproducibility statement}
The source code of our proposed approach is available at \url{https://github.com/KPeng9510/ERELIFM} to ensure reproducibility.

\section*{Ethics statement}
This work presents a methodological contribution to open-set domain generalization under noisy labels and is conducted entirely on publicly available benchmark datasets that do not involve human subjects or sensitive personal information. The research does not raise concerns regarding privacy, security, fairness, bias, or potential harmful applications, and complies with accepted standards of research integrity and ethical practice.
\bibliography{iclr2026_conference}

\begin{thebibliography}{77}
\providecommand{\natexlab}[1]{#1}
\providecommand{\url}[1]{\texttt{#1}}
\expandafter\ifx\csname urlstyle\endcsname\relax
  \providecommand{\doi}[1]{doi: #1}\else
  \providecommand{\doi}{doi: \begingroup \urlstyle{rm}\Url}\fi

\bibitem[Bao et~al.(2021)Bao, Yu, and Kong]{bao2021evidential}
Wentao Bao, Qi~Yu, and Yu~Kong.
\newblock Evidential deep learning for open set action recognition.
\newblock In \emph{ICCV}, 2021.

\bibitem[Beery et~al.(2018)Beery, Van~Horn, and Perona]{beery2018recognition}
Sara Beery, Grant Van~Horn, and Pietro Perona.
\newblock Recognition in terra incognita.
\newblock In \emph{ECCV}, 2018.

\bibitem[Bendale \& Boult(2016)Bendale and Boult]{bendale2016towards}
Abhijit Bendale and Terrance~E. Boult.
\newblock Towards open set deep networks.
\newblock In \emph{CVPR}, 2016.

\bibitem[Bose et~al.(2023)Bose, Jha, Kandala, and Banerjee]{bose2023beyond}
Shirsha Bose, Ankit Jha, Hitesh Kandala, and Biplab Banerjee.
\newblock Beyond boundaries: A novel data-augmentation discourse for open domain generalization.
\newblock \emph{TMLR}, 2023.

\bibitem[Busto et~al.(2020)Busto, Iqbal, and Gall]{busto2018open}
Pau~Panareda Busto, Ahsan Iqbal, and Juergen Gall.
\newblock Open set domain adaptation for image and action recognition.
\newblock \emph{TPAMI}, 2020.

\bibitem[Cha et~al.(2021)Cha, Chun, Lee, Cho, Park, Lee, and Park]{cha2021swad}
Junbum Cha, Sanghyuk Chun, Kyungjae Lee, Han-Cheol Cho, Seunghyun Park, Yunsung Lee, and Sungrae Park.
\newblock Swad: Domain generalization by seeking flat minima.
\newblock In \emph{NeurIPS}, 2021.

\bibitem[Chen et~al.(2022)Chen, Peng, Wang, and Tian]{chen2021adversarial}
Guangyao Chen, Peixi Peng, Xiangqian Wang, and Yonghong Tian.
\newblock Adversarial reciprocal points learning for open set recognition.
\newblock \emph{TPAMI}, 2022.

\bibitem[Chen et~al.(2021)Chen, Ye, Chen, Zhao, and Heng]{chen2021beyond}
Pengfei Chen, Junjie Ye, Guangyong Chen, Jingwei Zhao, and Pheng-Ann Heng.
\newblock Beyond class-conditional assumption: A primary attempt to combat instance-dependent label noise.
\newblock In \emph{AAAI}, 2021.

\bibitem[Chen \& Lipman(2023)Chen and Lipman]{chen2023riemannian}
Ricky T.~Q. Chen and Yaron Lipman.
\newblock Riemannian flow matching on general geometries.
\newblock \emph{arXiv preprint arXiv:2302.03660}, 2023.

\bibitem[Cheng et~al.(2023)Cheng, Zhu, Sun, and Liu]{cheng2021mitigating}
Hao Cheng, Zhaowei Zhu, Xing Sun, and Yang Liu.
\newblock Mitigating memorization of noisy labels via regularization between representations.
\newblock In \emph{ICLR}, 2023.

\bibitem[Cheng et~al.(2020)Cheng, Zhou, Zhao, Li, Shang, Zheng, Pan, and Xu]{cheng2020weakly}
Lele Cheng, Xiangzeng Zhou, Liming Zhao, Dangwei Li, Hong Shang, Yun Zheng, Pan Pan, and Yinghui Xu.
\newblock Weakly supervised learning with side information for noisy labeled images.
\newblock In \emph{ECCV}, 2020.

\bibitem[Dao et~al.(2023)Dao, Phung, Nguyen, and Tran]{dao2023flow}
Quan Dao, Hao Phung, Binh Nguyen, and Anh Tran.
\newblock Flow matching in latent space.
\newblock \emph{arXiv preprint arXiv:2307.08698}, 2023.

\bibitem[Devlin et~al.(2019)Devlin, Chang, Lee, and Toutanova]{BERT}
Jacob Devlin, Ming{-}Wei Chang, Kenton Lee, and Kristina Toutanova.
\newblock {BERT:} {Pre-training} of deep bidirectional transformers for language understanding.
\newblock In Jill Burstein, Christy Doran, and Thamar Solorio (eds.), \emph{NAACL-HLT}, 2019.

\bibitem[Dosovitskiy et~al.(2021)Dosovitskiy, Beyer, Kolesnikov, Weissenborn, Zhai, Unterthiner, Dehghani, Minderer, Heigold, Gelly, Uszkoreit, and Houlsby]{dosovitskiy2021an}
Alexey Dosovitskiy, Lucas Beyer, Alexander Kolesnikov, Dirk Weissenborn, Xiaohua Zhai, Thomas Unterthiner, Mostafa Dehghani, Matthias Minderer, Georg Heigold, Sylvain Gelly, Jakob Uszkoreit, and Neil Houlsby.
\newblock An image is worth 16x16 words: Transformers for image recognition at scale.
\newblock In \emph{ICLR}, 2021.

\bibitem[Eijkelboom et~al.(2024)Eijkelboom, Bartosh, Andersson~Naesseth, Welling, and van~de Meent]{eijkelboom2024variational}
Floor Eijkelboom, Grigory Bartosh, Christian Andersson~Naesseth, Max Welling, and Jan-Willem van~de Meent.
\newblock Variational flow matching for graph generation.
\newblock In \emph{NeurIPS}, 2024.

\bibitem[Fu et~al.(2020)Fu, Cao, Long, and Wang]{fu2020learning}
Bo~Fu, Zhangjie Cao, Mingsheng Long, and Jianmin Wang.
\newblock Learning to detect open classes for universal domain adaptation.
\newblock In \emph{ECCV}, 2020.

\bibitem[Gat et~al.(2024)Gat, Remez, Shaul, Kreuk, Chen, Synnaeve, Adi, and Lipman]{gat2024discrete}
Itai Gat, Tal Remez, Neta Shaul, Felix Kreuk, Ricky T.~Q. Chen, Gabriel Synnaeve, Yossi Adi, and Yaron Lipman.
\newblock Discrete flow matching.
\newblock In \emph{NeurIPS}, 2024.

\bibitem[Geng et~al.(2021)Geng, Huang, and Chen]{geng2020recent}
Chuanxing Geng, Sheng-jun Huang, and Songcan Chen.
\newblock Recent advances in open set recognition: A survey.
\newblock \emph{TPAMI}, 2021.

\bibitem[Guo et~al.(2023)Guo, Wang, Qi, and Shi]{guo2023aloft}
Jintao Guo, Na~Wang, Lei Qi, and Yinghuan Shi.
\newblock {ALOFT:} {A} lightweight {MLP-like} architecture with dynamic low-frequency transform for domain generalization.
\newblock In \emph{CVPR}, 2023.

\bibitem[Guo et~al.(2022)Guo, Liu, Liu, and Yuan]{guo2022simt}
Xiaoqing Guo, Jie Liu, Tongliang Liu, and Yixuan Yuan.
\newblock {SimT:} {Handling} open-set noise for domain adaptive semantic segmentation.
\newblock In \emph{CVPR}, 2022.

\bibitem[Han et~al.(2018)Han, Yao, Yu, Niu, Xu, Hu, Tsang, and Sugiyama]{han2018co}
Bo~Han, Quanming Yao, Xingrui Yu, Gang Niu, Miao Xu, Weihua Hu, Ivor Tsang, and Masashi Sugiyama.
\newblock Co-teaching: Robust training of deep neural networks with extremely noisy labels.
\newblock \emph{NeurIPS}, 2018.

\bibitem[He et~al.(2016)He, Zhang, Ren, and Sun]{he2016deep}
Kaiming He, Xiangyu Zhang, Shaoqing Ren, and Jian Sun.
\newblock Deep residual learning for image recognition.
\newblock In \emph{CVPR}, 2016.

\bibitem[Ho et~al.(2020)Ho, Jain, and Abbeel]{Ho2020DenoisingDP}
Jonathan Ho, Ajay Jain, and Pieter Abbeel.
\newblock Denoising diffusion probabilistic models.
\newblock In \emph{NeurIPS}, 2020.

\bibitem[Huang et~al.(2023)Huang, Zhang, and Shan]{huang2023twin}
Zhizhong Huang, Junping Zhang, and Hongming Shan.
\newblock Twin contrastive learning with noisy labels.
\newblock In \emph{CVPR}, 2023.

\bibitem[Karim et~al.(2022)Karim, Rizve, Rahnavard, Mian, and Shah]{karim2022unicon}
Nazmul Karim, Mamshad~Nayeem Rizve, Nazanin Rahnavard, Ajmal Mian, and Mubarak Shah.
\newblock {UNICON:} {Combating} label noise through uniform selection and contrastive learning.
\newblock In \emph{CVPR}, 2022.

\bibitem[Katsumata et~al.(2021)Katsumata, Kishida, Amma, and Nakayama]{katsumata2021open}
Kai Katsumata, Ikki Kishida, Ayako Amma, and Hideki Nakayama.
\newblock Open-set domain generalization via metric learning.
\newblock In \emph{ICIP}, 2021.

\bibitem[Kim et~al.(2024)Kim, Lee, and Lee]{kim2024learning}
Noo-ri Kim, Jin-Seop Lee, and Jee-Hyong Lee.
\newblock Learning with structural labels for learning with noisy labels.
\newblock In \emph{CVPR}, 2024.

\bibitem[Kim et~al.(2021)Kim, Ko, Cho, Choi, and Yun]{kim2021fine}
Taehyeon Kim, Jongwoo Ko, Sangwook Cho, Jinhwan Choi, and Se{-}Young Yun.
\newblock {FINE} samples for learning with noisy labels.
\newblock In \emph{NeurIPS}, 2021.

\bibitem[Klein et~al.(2023)Klein, Kr{\"a}mer, and No{\'e}]{klein2023equivariant}
Leon Klein, Andreas Kr{\"a}mer, and Frank No{\'e}.
\newblock Equivariant flow matching.
\newblock In \emph{NeurIPS}, 2023.

\bibitem[Li et~al.(2017)Li, Yang, Song, and Hospedales]{li2017deeper}
Da~Li, Yongxin Yang, Yi{-}Zhe Song, and Timothy~M. Hospedales.
\newblock Deeper, broader and artier domain generalization.
\newblock In \emph{ICCV}, 2017.

\bibitem[Li et~al.(2020)Li, Wang, Wan, Wang, Li, and Kot]{li2020domain}
Haoliang Li, YuFei Wang, Renjie Wan, Shiqi Wang, Tie-Qiang Li, and Alex Kot.
\newblock Domain generalization for medical imaging classification with linear-dependency regularization.
\newblock In \emph{NeurIPS}, 2020.

\bibitem[Li et~al.(2021{\natexlab{a}})Li, Gao, Cao, Huang, Weng, Mi, Yu, Li, and Xia]{li2021progressive}
Lei Li, Ke~Gao, Juan Cao, Ziyao Huang, Yepeng Weng, Xiaoyue Mi, Zhengze Yu, Xiaoya Li, and Boyang Xia.
\newblock Progressive domain expansion network for single domain generalization.
\newblock In \emph{CVPR}, 2021{\natexlab{a}}.

\bibitem[Li et~al.(2024)Li, Xiao, Shi, Zhang, Yang, Liu, and Chen]{li2024nicest}
Lin Li, Jun Xiao, Hanrong Shi, Hanwang Zhang, Yi~Yang, Wei Liu, and Long Chen.
\newblock {NICEST:} {Noisy} label correction and training for robust scene graph generation.
\newblock \emph{TPAMI}, 2024.

\bibitem[Li et~al.(2021{\natexlab{b}})Li, Li, Li, Gong, Fu, and Hospedales]{li2021simple}
Pan Li, Da~Li, Wei Li, Shaogang Gong, Yanwei Fu, and Timothy~M. Hospedales.
\newblock A simple feature augmentation for domain generalization.
\newblock In \emph{ICCV}, 2021{\natexlab{b}}.

\bibitem[Li et~al.(2022)Li, Xia, Zhang, Zhan, Ge, and Liu]{li2022estimating}
Shikun Li, Xiaobo Xia, Hansong Zhang, Yibing Zhan, Shiming Ge, and Tongliang Liu.
\newblock Estimating noise transition matrix with label correlations for noisy multi-label learning.
\newblock In \emph{NeurIPS}, 2022.

\bibitem[Li et~al.(2018)Li, Tian, Gong, Liu, Liu, Zhang, and Tao]{li2018deep}
Ya~Li, Xinmei Tian, Mingming Gong, Yajing Liu, Tongliang Liu, Kun Zhang, and Dacheng Tao.
\newblock Deep domain generalization via conditional invariant adversarial networks.
\newblock In \emph{ECCV}, 2018.

\bibitem[Li et~al.(2023)Li, Han, Shan, and Chen]{li2023disc}
Yifan Li, Hu~Han, Shiguang Shan, and Xilin Chen.
\newblock {DISC:} {Learning} from noisy labels via dynamic instance-specific selection and correction.
\newblock In \emph{CVPR}, 2023.

\bibitem[Lipman et~al.(2023)Lipman, Chen, Ben{-}Hamu, Nickel, and Le]{lipman2022flow}
Yaron Lipman, Ricky T.~Q. Chen, Heli Ben{-}Hamu, Maximilian Nickel, and Matthew Le.
\newblock Flow matching for generative modeling.
\newblock In \emph{ICLR}, 2023.

\bibitem[Liu et~al.(2021)Liu, Li, and Sun]{liu2021co}
Jiarun Liu, Ruirui Li, and Chuan Sun.
\newblock Co-correcting: noise-tolerant medical image classification via mutual label correction.
\newblock \emph{TMI}, 2021.

\bibitem[Liu et~al.(2022)Liu, Zhu, Qu, and You]{liu2022robust}
Sheng Liu, Zhihui Zhu, Qing Qu, and Chong You.
\newblock Robust training under label noise by over-parameterization.
\newblock In \emph{ICML}, 2022.

\bibitem[Liu \& Tao(2016)Liu and Tao]{liu2015classification}
Tongliang Liu and Dacheng Tao.
\newblock Classification with noisy labels by importance reweighting.
\newblock \emph{TPAMI}, 2016.

\bibitem[Liu et~al.(2023)Liu, Gong, and Liu]{liu2022flow}
Xingchao Liu, Chengyue Gong, and Qiang Liu.
\newblock Flow straight and fast: Learning to generate and transfer data with rectified flow.
\newblock In \emph{ICLR}, 2023.

\bibitem[Liu \& Guo(2020)Liu and Guo]{liu2020peer}
Yang Liu and Hongyi Guo.
\newblock Peer loss functions: Learning from noisy labels without knowing noise rates.
\newblock In \emph{ICML}, 2020.

\bibitem[Ma et~al.(2020)Ma, Huang, Wang, Romano, Erfani, and Bailey]{ma2020normalized}
Xingjun Ma, Hanxun Huang, Yisen Wang, Simone Romano, Sarah Erfani, and James Bailey.
\newblock Normalized loss functions for deep learning with noisy labels.
\newblock In \emph{ICML}, 2020.

\bibitem[Nam et~al.(2021)Nam, Lee, Park, Yoon, and Yoo]{nam2021reducing}
Hyeonseob Nam, HyunJae Lee, Jongchan Park, Wonjun Yoon, and Donggeun Yoo.
\newblock Reducing domain gap by reducing style bias.
\newblock In \emph{CVPR}, 2021.

\bibitem[Peebles \& Xie(2023)Peebles and Xie]{Peebles2022DiT}
William Peebles and Saining Xie.
\newblock Scalable diffusion models with transformers.
\newblock In \emph{ICCV}, 2023.

\bibitem[Peng et~al.(2024{\natexlab{a}})Peng, Wen, Yang, Luo, Chen, Fu, Sarfraz, Roitberg, and Stiefelhagen]{peng2024advancing}
Kunyu Peng, Di~Wen, Kailun Yang, Ao~Luo, Yufan Chen, Jia Fu, M.~Saquib Sarfraz, Alina Roitberg, and Rainer Stiefelhagen.
\newblock Advancing open-set domain generalization using evidential bi-level hardest domain scheduler.
\newblock In \emph{NeurIPS}, 2024{\natexlab{a}}.

\bibitem[Peng et~al.(2024{\natexlab{b}})Peng, Yin, Zheng, Liu, Schneider, Zhang, Yang, Sarfraz, Stiefelhagen, and Roitberg]{peng2024navigating}
Kunyu Peng, Cheng Yin, Junwei Zheng, Ruiping Liu, David Schneider, Jiaming Zhang, Kailun Yang, M.~Saquib Sarfraz, Rainer Stiefelhagen, and Alina Roitberg.
\newblock Navigating open set scenarios for skeleton-based action recognition.
\newblock In \emph{AAAI}, 2024{\natexlab{b}}.

\bibitem[Peng et~al.(2025)Peng, Wen, Saquib, Chen, Zheng, Schneider, Yang, Wu, Roitberg, and Stiefelhagen]{peng2024mitigating}
Kunyu Peng, Di~Wen, Sarfraz~M. Saquib, Yufan Chen, Junwei Zheng, David Schneider, Kailun Yang, Jiamin Wu, Alina Roitberg, and Rainer Stiefelhagen.
\newblock Mitigating label noise using prompt-based hyperbolic meta-learning in open-set domain generalization.
\newblock \emph{IJCV}, 2025.

\bibitem[Peng et~al.(2019)Peng, Bai, Xia, Huang, Saenko, and Wang]{peng2019moment}
Xingchao Peng, Qinxun Bai, Xide Xia, Zijun Huang, Kate Saenko, and Bo~Wang.
\newblock Moment matching for multi-source domain adaptation.
\newblock In \emph{ICCV}, 2019.

\bibitem[Sarfraz et~al.(2019)Sarfraz, Sharma, and Stiefelhagen]{finch}
M.~Saquib Sarfraz, Vivek Sharma, and Rainer Stiefelhagen.
\newblock Efficient parameter-free clustering using first neighbor relations.
\newblock In \emph{CVPR}, 2019.

\bibitem[Sheng et~al.(2024)Sheng, Sun, Cai, Chen, Zhou, and Yao]{sheng2024adaptive}
Mengmeng Sheng, Zeren Sun, Zhenhuang Cai, Tao Chen, Yichao Zhou, and Yazhou Yao.
\newblock Adaptive integration of partial label learning and negative learning for enhanced noisy label learning.
\newblock In \emph{AAAI}, 2024.

\bibitem[Shu et~al.(2019)Shu, Xie, Yi, Zhao, Zhou, Xu, and Meng]{shu2019meta}
Jun Shu, Qi~Xie, Lixuan Yi, Qian Zhao, Sanping Zhou, Zongben Xu, and Deyu Meng.
\newblock {Meta-Weight-Net:} {Learning} an explicit mapping for sample weighting.
\newblock In \emph{NeurIPS}, 2019.

\bibitem[Shu et~al.(2021)Shu, Cao, Wang, Wang, and Long]{shu2021open}
Yang Shu, Zhangjie Cao, Chenyu Wang, Jianmin Wang, and Mingsheng Long.
\newblock Open domain generalization with domain-augmented meta-learning.
\newblock In \emph{CVPR}, 2021.

\bibitem[Singha et~al.(2024)Singha, Jha, Bose, Nair, Abdar, and Banerjee]{singha2024unknown}
Mainak Singha, Ankit Jha, Shirsha Bose, Ashwin Nair, Moloud Abdar, and Biplab Banerjee.
\newblock Unknown prompt, the only lacuna: Unveiling {CLIP's} potential for open domain generalization.
\newblock \emph{arXiv preprint arXiv:2404.00710}, 2024.

\bibitem[Song et~al.(2019)Song, Kim, and Lee]{song2019selfie}
Hwanjun Song, Minseok Kim, and Jae-Gil Lee.
\newblock {SELFIE:} {Refurbishing} unclean samples for robust deep learning.
\newblock In \emph{ICML}, 2019.

\bibitem[Tanno et~al.(2019)Tanno, Saeedi, Sankaranarayanan, Alexander, and Silberman]{tanno2019learning}
Ryutaro Tanno, Ardavan Saeedi, Swami Sankaranarayanan, Daniel~C. Alexander, and Nathan Silberman.
\newblock Learning from noisy labels by regularized estimation of annotator confusion.
\newblock In \emph{CVPR}, 2019.

\bibitem[Wang et~al.(2024)Wang, Zhao, Zhang, and Feng]{wang2024towards}
Ruofan Wang, Rui-Wei Zhao, Xiaobo Zhang, and Rui Feng.
\newblock Towards evidential and class separable open set object detection.
\newblock In \emph{AAAI}, 2024.

\bibitem[Wang et~al.(2023)Wang, Zhang, Qi, and Shi]{wang2023generalizable}
Xiran Wang, Jian Zhang, Lei Qi, and Yinghuan Shi.
\newblock Generalizable decision boundaries: Dualistic meta-learning for open set domain generalization.
\newblock In \emph{ICCV}, 2023.

\bibitem[Wang et~al.(2020)Wang, Li, and Kot]{wang2020heterogeneous}
Yufei Wang, Haoliang Li, and Alex~C. Kot.
\newblock Heterogeneous domain generalization via domain mixup.
\newblock In \emph{ICASSP}, 2020.

\bibitem[Wei et~al.(2021)Wei, Tao, Xie, and An]{wei2021open}
Hongxin Wei, Lue Tao, Renchunzi Xie, and Bo~An.
\newblock Open-set label noise can improve robustness against inherent label noise.
\newblock In \emph{NeurIPS}, 2021.

\bibitem[Wei et~al.(2022)Wei, Liu, Liu, Niu, Sugiyama, and Liu]{wei2021smooth}
Jiaheng Wei, Hangyu Liu, Tongliang Liu, Gang Niu, Masashi Sugiyama, and Yang Liu.
\newblock To smooth or not? {When} label smoothing meets noisy labels.
\newblock In \emph{ICML}, 2022.

\bibitem[Xia et~al.(2019)Xia, Liu, Wang, Han, Gong, Niu, and Sugiyama]{xia2019anchor}
Xiaobo Xia, Tongliang Liu, Nannan Wang, Bo~Han, Chen Gong, Gang Niu, and Masashi Sugiyama.
\newblock Are anchor points really indispensable in label-noise learning?
\newblock In \emph{NeurIPS}, 2019.

\bibitem[Xu et~al.(2024)Xu, Peng, Wen, Liu, Zheng, Chen, Zhang, Roitberg, Yang, and Stiefelhagen]{xu2024skeleton}
Yi~Xu, Kunyu Peng, Di~Wen, Ruiping Liu, Junwei Zheng, Yufan Chen, Jiaming Zhang, Alina Roitberg, Kailun Yang, and Rainer Stiefelhagen.
\newblock Skeleton-based human action recognition with noisy labels.
\newblock In \emph{IROS}, 2024.

\bibitem[Yue \& Jha(2024)Yue and Jha]{yue2024ctrl}
Chang Yue and Niraj~K. Jha.
\newblock {CTRL:} {Clustering} training losses for label error detection.
\newblock \emph{TAI}, 2024.

\bibitem[Zhang et~al.(2024)Zhang, Song, Wang, Han, Liu, Liu, and Sugiyama]{zhang2024badlabel}
Jingfeng Zhang, Bo~Song, Haohan Wang, Bo~Han, Tongliang Liu, Lei Liu, and Masashi Sugiyama.
\newblock {BadLabel:} {A} robust perspective on evaluating and enhancing label-noise learning.
\newblock \emph{TPAMI}, 2024.

\bibitem[Zhao \& Shen(2022)Zhao and Shen]{zhao2022adaptive}
Chao Zhao and Weiming Shen.
\newblock Adaptive open set domain generalization network: Learning to diagnose unknown faults under unknown working conditions.
\newblock \emph{Reliability Engineering \& System Safety}, 2022.

\bibitem[Zhao et~al.(2023)Zhao, Du, Hoogs, and Funk]{zhao2023open}
Chen Zhao, Dawei Du, Anthony Hoogs, and Christopher Funk.
\newblock Open set action recognition via multi-label evidential learning.
\newblock In \emph{CVPR}, 2023.

\bibitem[Zhao et~al.(2024)Zhao, Shi, Ruan, Pan, and Dong]{zhao2024estimating}
Rui Zhao, Bin Shi, Jianfei Ruan, Tianze Pan, and Bo~Dong.
\newblock Estimating noisy class posterior with part-level labels for noisy label learning.
\newblock In \emph{CVPR}, 2024.

\bibitem[Zheng et~al.(2020)Zheng, Wu, Goswami, Goswami, Metaxas, and Chen]{zheng2020error}
Songzhu Zheng, Pengxiang Wu, Aman Goswami, Mayank Goswami, Dimitris Metaxas, and Chao Chen.
\newblock Error-bounded correction of noisy labels.
\newblock In \emph{ICML}, 2020.

\bibitem[Zhou et~al.(2020{\natexlab{a}})Zhou, Yang, Hospedales, and Xiang]{zhou2020deep}
Kaiyang Zhou, Yongxin Yang, Timothy Hospedales, and Tao Xiang.
\newblock Deep domain-adversarial image generation for domain generalisation.
\newblock In \emph{AAAI}, 2020{\natexlab{a}}.

\bibitem[Zhou et~al.(2020{\natexlab{b}})Zhou, Yang, Hospedales, and Xiang]{zhou2020learning}
Kaiyang Zhou, Yongxin Yang, Timothy Hospedales, and Tao Xiang.
\newblock Learning to generate novel domains for domain generalization.
\newblock In \emph{ECCV}, 2020{\natexlab{b}}.

\bibitem[Zhou et~al.(2020{\natexlab{c}})Zhou, Yang, Qiao, and Xiang]{zhou2020domain}
Kaiyang Zhou, Yongxin Yang, Yu~Qiao, and Tao Xiang.
\newblock Domain generalization with mixstyle.
\newblock In \emph{ICLR}, 2020{\natexlab{c}}.

\bibitem[Zhou et~al.(2021)Zhou, Yang, Qiao, and Xiang]{zhou2021domain}
Kaiyang Zhou, Yongxin Yang, Yu~Qiao, and Tao Xiang.
\newblock Domain adaptive ensemble learning.
\newblock \emph{TIP}, 2021.

\bibitem[Zhu et~al.(2021{\natexlab{a}})Zhu, Liu, and Liu]{zhu2021second}
Zhaowei Zhu, Tongliang Liu, and Yang Liu.
\newblock A second-order approach to learning with instance-dependent label noise.
\newblock In \emph{CVPR}, 2021{\natexlab{a}}.

\bibitem[Zhu et~al.(2021{\natexlab{b}})Zhu, Song, and Liu]{zhu2021clusterability}
Zhaowei Zhu, Yiwen Song, and Yang Liu.
\newblock Clusterability as an alternative to anchor points when learning with noisy labels.
\newblock In \emph{ICML}, 2021{\natexlab{b}}.

\bibitem[Zhu et~al.(2022)Zhu, Wang, and Liu]{zhu2022beyond}
Zhaowei Zhu, Jialu Wang, and Yang Liu.
\newblock Beyond images: Label noise transition matrix estimation for tasks with lower-quality features.
\newblock In \emph{ICML}, 2022.

\end{thebibliography}
\bibliographystyle{iclr2026_conference}
\clearpage

\appendix
\section*{Appendix}
\section{The Use of Large Language Models (LLMs)
}
In this work, we mainly rely on LLM for text rephrasing to polish the paper writing.
\section{Social Impact and Limitations}
\noindent\textbf{Social impact:} The proposed EReLiFM framework has a significant social impact by improving model generalization to new categories and domains under noisy labels, which is critical for real-world applications such as healthcare, security, and autonomous driving. By facilitating robust learning in open-set environments, this work enhances the reliability of deep learning models deployed in dynamic and uncertain conditions when label noise exist. The ability to manage label noise ensures that models trained on imperfect annotations, such as crowdsourced data, maintain their effectiveness and trustworthiness. Furthermore, our approach mitigates biases in deep learning-based decision-making by distinguishing between reliable and noisy labels, contributing to fairer and more accountable deep learning systems. However, the potential misclassification and biased prediction remain, which could lead to erroneous decisions with adverse societal implications.

\noindent\textbf{Limitations:} We propose EReLiFM to mitigate label noise in OSDG, but its performance under extreme noise remains limited, highlighting a key research direction. This work focuses on image-based OSDG-NL, leaving video-based OSDG-NL for future exploration.

\section{More Details regarding the Evaluation Metrics}
We follow the protocol outlined in the MEDIC approach~\cite{wang2023generalizable}. For the PACS~\cite{li2017deeper} dataset, we adopt an open-set ratio of $6:1$, designating \textit{elephant}, \textit{horse}, \textit{giraffe}, \textit{dog}, \textit{guitar}, and \textit{house} as seen categories, while \textit{person} is treated as unseen. Similarly, in DigitsDG~\cite{zhou2020deep}, we use an open-set ratio of $6:4$, with digits $0,1,2,3,4,5$ as seen and $6,7,8,9$ as unseen.

For evaluation, we employ three metrics. \textit{Acc} measures closed-set accuracy on seen categories, while \textit{H-score} and \textit{OSCR} assess open-set recognition. The \textit{H-score}, dependent on a threshold from the source domain validation set, is considered as the secondary metric. In contrast, OSCR, introduced by MEDIC~\cite{wang2023generalizable}, evaluates open-set recognition without a predefined threshold, making it our primary metric.

The \textit{H-score} is computed using a threshold ratio $\lambda$ to distinguish seen from unseen samples. Predictions below $\lambda$ are classified as unseen, and accuracy is separately calculated for seen ($Acc_k$) and unseen ($Acc_u$) categories. The final \textit{H-score} is given by:

\begin{equation} H_{score} = \frac{2 \times Acc_u \times Acc_k}{Acc_u + Acc_k}. \end{equation}

OSCR, unlike AUROC, integrates accuracy with AUROC through dynamic thresholding, focusing only on correctly classified samples. It combines elements from both \textit{H-score} and AUROC, offering a more comprehensive measure of confidence reliability in OSDG tasks.
\section{Analysis of Confidence Score}
\begin{figure*}[htb]
\centering
\begin{subfigure}[b]{0.19\textwidth}
    \centering
    \includegraphics[width=\textwidth]{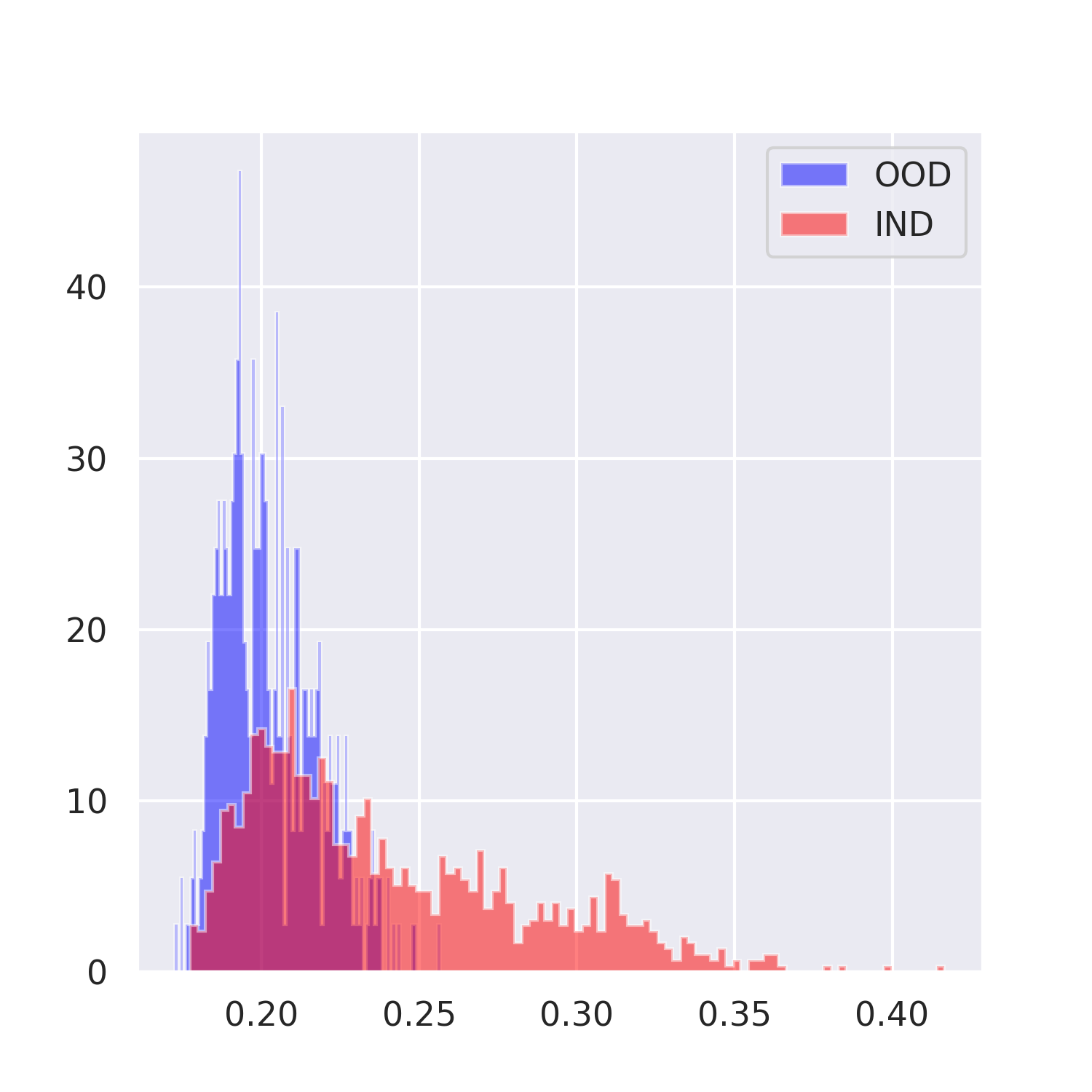}
        %\vskip-2ex
    \caption{BadLabel}
    \label{fig:bad_label_photo}
\end{subfigure}
\hfill
\begin{subfigure}[b]{0.19\textwidth}
    \centering
    \includegraphics[width=\textwidth]{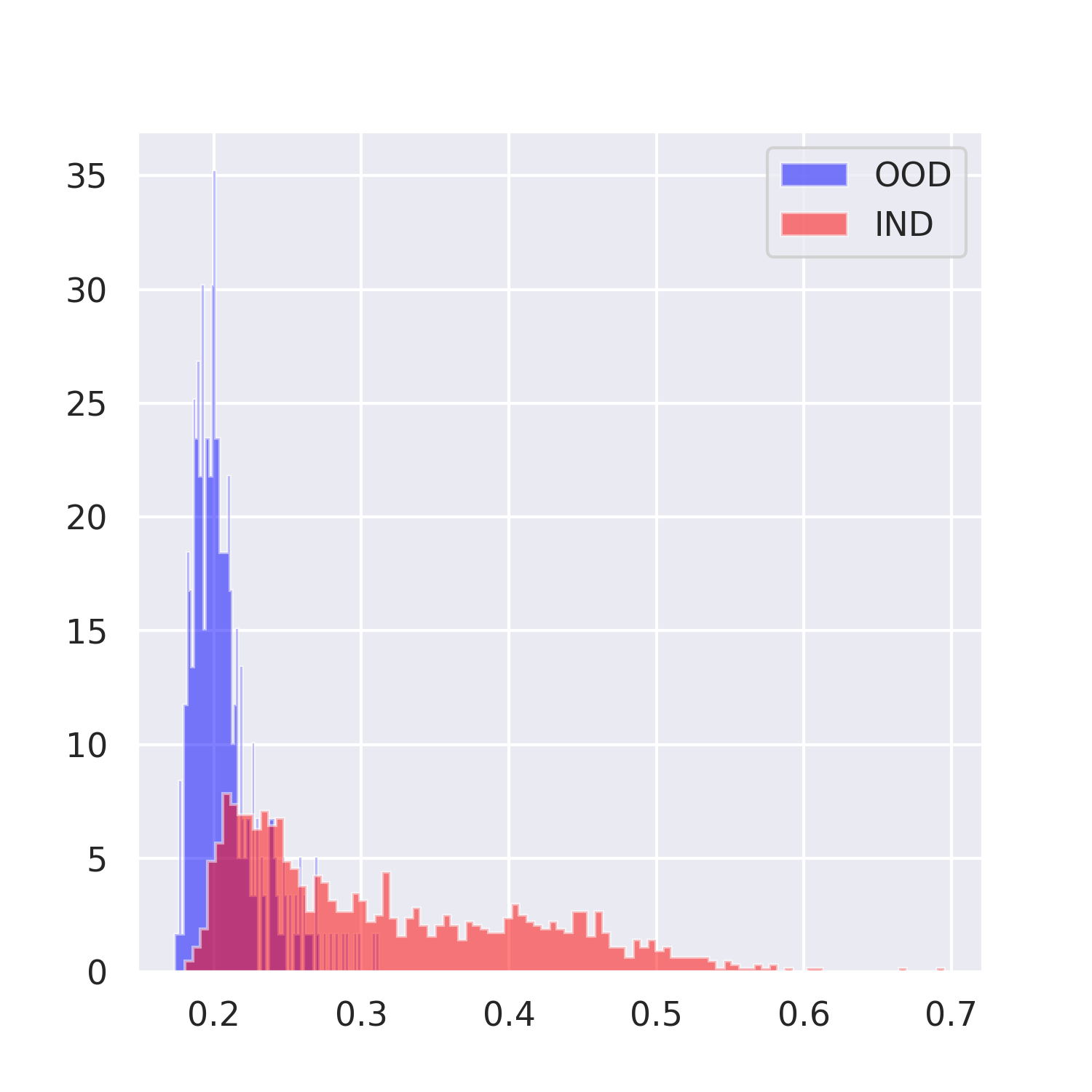}
        %\vskip-2ex
    \caption{MLDG}
    \label{fig:MLDG_photo}
\end{subfigure}
\hfill
\begin{subfigure}[b]{0,19\textwidth}
    \centering
    \includegraphics[width=\textwidth]{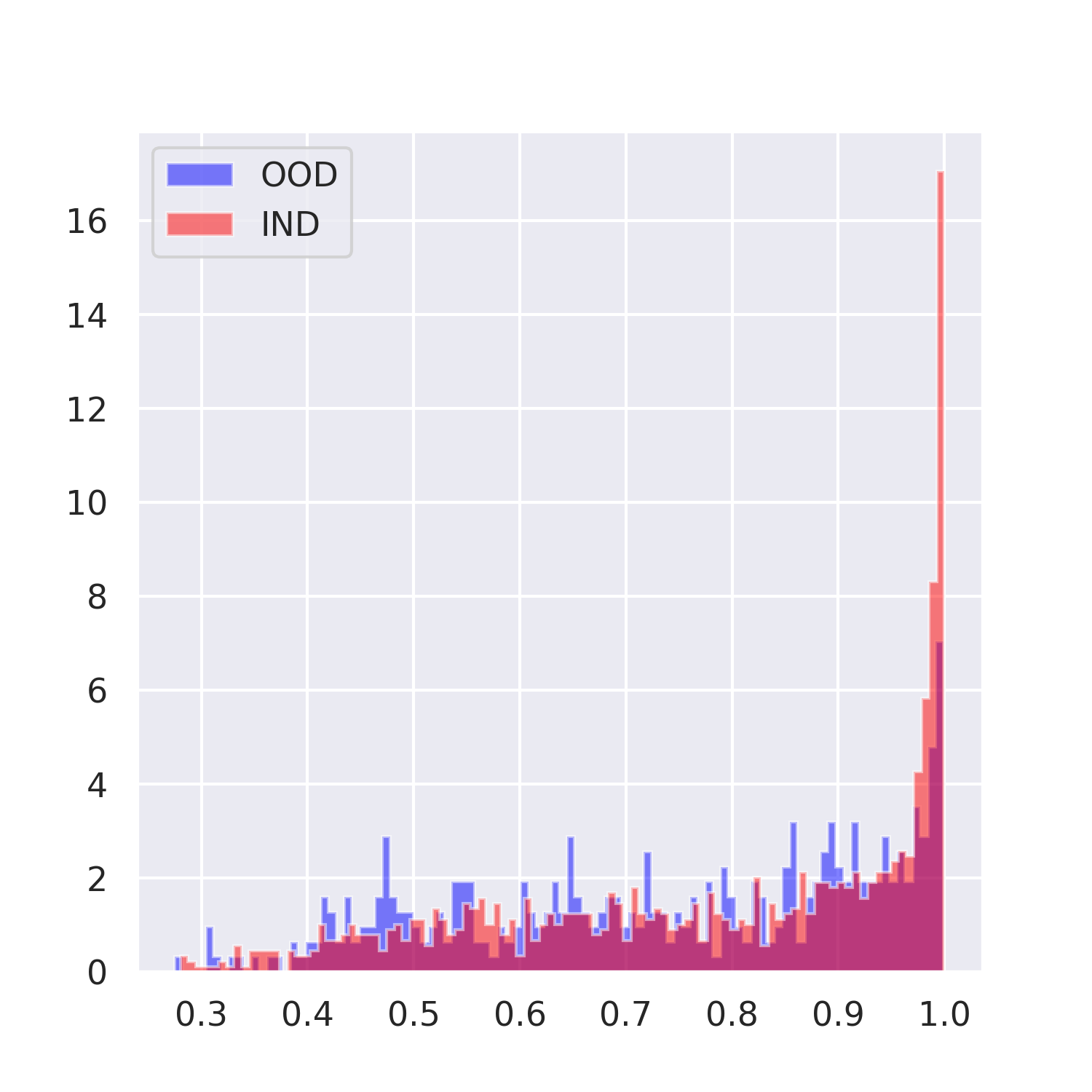}
        %\vskip-2ex
    \caption{MEDIC}
    \label{fig:medic_photo}
\end{subfigure}
\hfill
\begin{subfigure}[b]{0.19\textwidth}
    \centering
    \includegraphics[width=\textwidth]{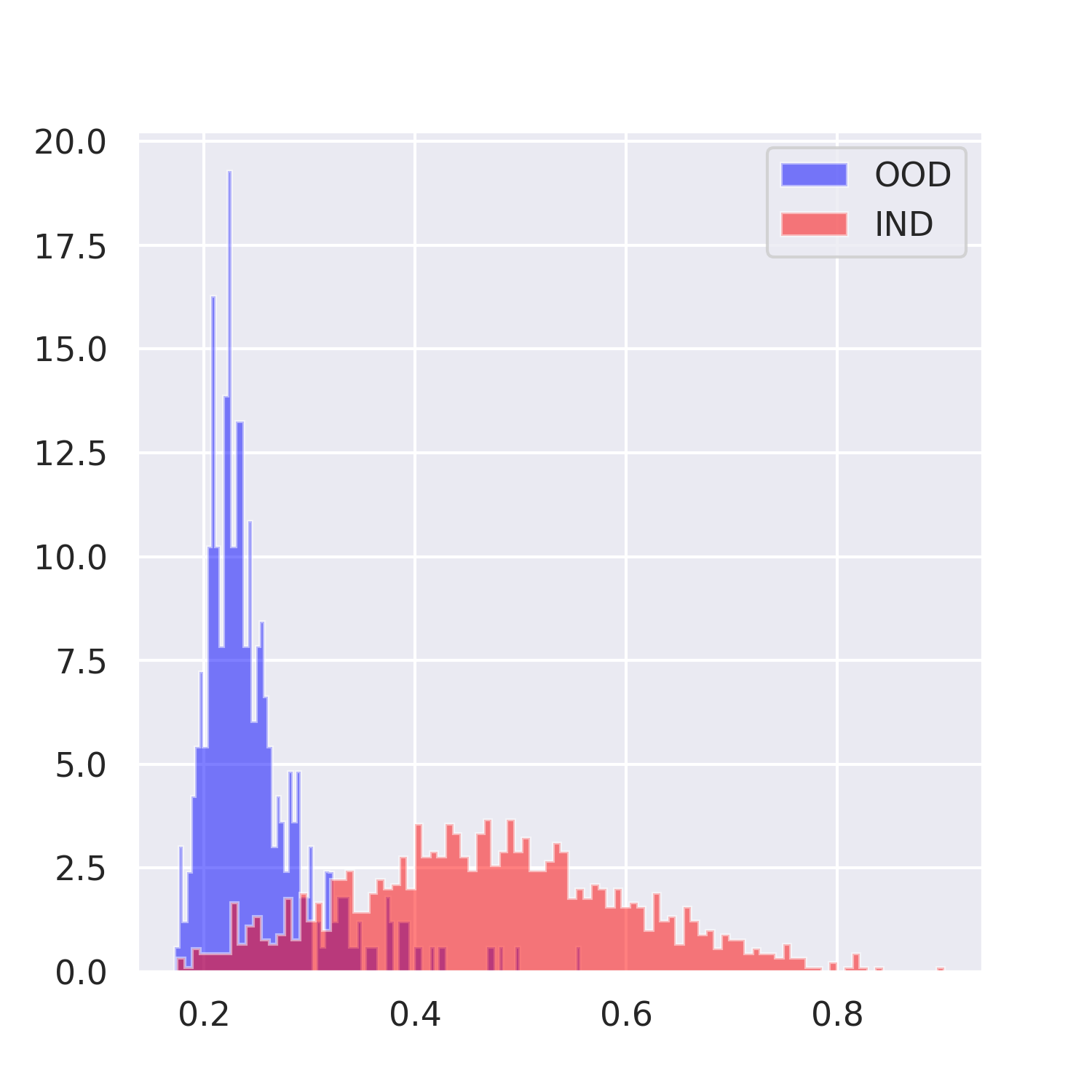}
        %\vskip-2ex
    \caption{HyProMeta}
    \label{fig:ours_photo}
\end{subfigure}
\hfill
\begin{subfigure}[b]{0.19\textwidth}
    \centering
    \includegraphics[width=\textwidth]{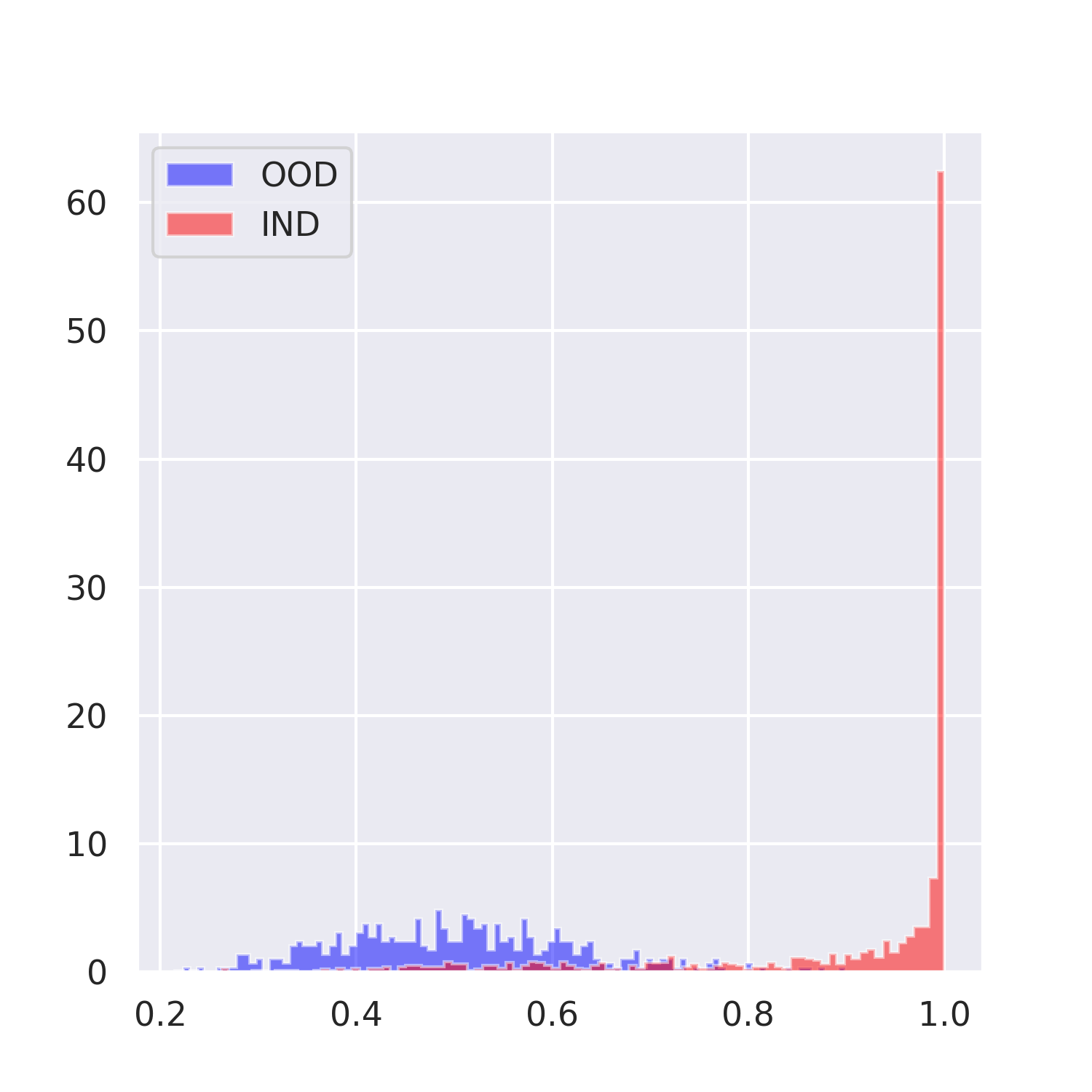}
        %\vskip-2ex
    \caption{Ours}
    \label{fig:bad_label_photo_a}
\end{subfigure}
\caption{Confidence score visualization of learned representations on PACS with target domain \textit{photo}, using ResNet18~\cite{he2016deep} under symmetric label noise with ratio $50\%$.}
\label{fig:conf_score}
\end{figure*}
Fig.~\ref{fig:conf_score} visualizes confidence scores for seen (red) and unseen (blue) categories, computed as the maximum Softmax probability. Our approach achieves the best separation between seen categories and unseen categories on the test domain, while the confidence scores delivered by other listed baselines are merged together. This visualization illustrates the superior capability of the proposed approach when it deals with out-of-distribution categories. The proposed categorical flow matching improves the awareness of unseen categories during the representation learning.

\section{Per-Target-Domain Results on PACS using ViT-Base and DigitsDG using ConvNet}

We further deliver the per-target-domain performances for the experiments conducted on the PACS~\cite{li2017deeper} dataset using ViT-Base~\cite{dosovitskiy2021an} backbone (as shown in Tab.~\ref{tab:pacs_vit_20_append}, Tab.~\ref{tab:pacs_vit_50}, Tab.~\ref{tab:pacs_vit_80}, and Tab.~\ref{tab:pacs_vit_50a}), and the experiments conducted on the DigitsDG~\cite{zhou2020deep} dataset using ConvNet~\cite{zhou2021domain} backbone (as shown in Tab.~\ref{tab:dg_20_append}, Tab.~\ref{tab:dg_50}, Tab.~\ref{tab:dg_80}, and Tab.~\ref{tab:dg_50a}). 
From the aforementioned tables, we can observe that our proposed approach consistently outperforms the others across all the metrics and label noise settings in general, which demonstrates the superior generalizability of our approach across different backbones, label noise settings, and datasets.

\begin{table*}[t!]

\centering
\resizebox{1.\linewidth}{!}{
\begin{tabular}{l|ccc|ccc|ccc|ccc|ccc}
\toprule
& \multicolumn{3}{c|}{\textbf{Photo (P)}} & \multicolumn{3}{c|}{\textbf{Art (A)}} & \multicolumn{3}{c|}{\textbf{Cartoon (C)}} & \multicolumn{3}{c|}{\textbf{Sketch (S)}} & \multicolumn{3}{c}{\textbf{Avg}} \\
\textbf{Method} & Acc & H-score & \cellcolor{gray!25}OSCR & Acc & H-score & \cellcolor{gray!25}OSCR & Acc & H-score & \cellcolor{gray!25}OSCR & Acc & H-score & \cellcolor{gray!25}OSCR & Acc & H-score & \cellcolor{gray!25}OSCR \\

\midrule

TCL~\cite{huang2023twin}  & 67.77 & 79.17 & \cellcolor{gray!25}67.64 & 64.35 & 63.45 & \cellcolor{gray!25}56.91 & 51.32 & 50.24 & \cellcolor{gray!25}42.30 & 26.43 & \textbf{28.25} & \cellcolor{gray!25}20.54 & 52.47 & 55.28 & \cellcolor{gray!25} 46.85 \\

NPN~\cite{sheng2024adaptive}  & 63.97 & 71.61 & \cellcolor{gray!25}62.53 & 57.22 & 51.43 & \cellcolor{gray!25}45.95 & 47.65 & 40.11 & \cellcolor{gray!25}32.85 & 21.86 & 13.01 & \cellcolor{gray!25}12.71 & 47.68 & 44.04 & \cellcolor{gray!25} 38.51 \\

BadLabel~\cite{zhang2024badlabel}  & 62.66 & 64.81 & \cellcolor{gray!25}57.87 & 51.59 & 60.49 & \cellcolor{gray!25}49.13 & 50.85 & 58.42 & \cellcolor{gray!25}47.75 & 31.15 & 18.54 & \cellcolor{gray!25}29.40 & 49.06 & 50.57 & \cellcolor{gray!25} 46.04 \\

DISC~\cite{li2023disc} & 66.69 & 71.98 & \cellcolor{gray!25}65.67 & 54.10 & 29.80 & \cellcolor{gray!25}41.07 & 53.48 & 41.72 & \cellcolor{gray!25}40.83 & 34.57 & 16.94 & \cellcolor{gray!25}24.16 & 52.21 & 40.11 & \cellcolor{gray!25} 42.93 \\

LSL~\cite{kim2024learning}  & 67.04 & 71.46 & \cellcolor{gray!25}63.82 & 63.10 & 63.42 & \cellcolor{gray!25}58.01 & 53.27 & 54.92 & \cellcolor{gray!25}47.14 & 28.44 & 4.37 & \cellcolor{gray!25}27.51 & 52.96 & 48.54 & \cellcolor{gray!25} 49.12 \\
PLM~\cite{zhao2024estimating}  & 65.73 & 61.01 & \cellcolor{gray!25}59.76 &65.73  & 61.01 & \cellcolor{gray!25}59.76 & 52.71 & 48.17 & \cellcolor{gray!25}46.32 & 27.57 & 17.94 & \cellcolor{gray!25}20.62 & 52.94 & 47.03 & \cellcolor{gray!25} 46.62 \\
\bottomrule
\end{tabular}

}
\caption{Results (\%) of PACS~\cite{li2017deeper} on ViT-Base~\cite{dosovitskiy2021an}.
The open-set ratio is $6{:}1$ and symmetric label noise with ratio $20\%$ is selected.}
\label{tab:pacs_vit_20_append}
%\vskip-2ex
\end{table*}

\begin{table*}[t!]

\centering
\resizebox{1.\linewidth}{!}{
\begin{tabular}{l|ccc|ccc|ccc|ccc|ccc}
\toprule
& \multicolumn{3}{c|}{\textbf{Photo (P)}} & \multicolumn{3}{c|}{\textbf{Art (A)}} & \multicolumn{3}{c|}{\textbf{Cartoon (C)}} & \multicolumn{3}{c|}{\textbf{Sketch (S)}} & \multicolumn{3}{c}{\textbf{Avg}} \\
\textbf{Method} & Acc & H-score & \cellcolor{gray!25}OSCR & Acc & H-score & \cellcolor{gray!25}OSCR & Acc & H-score & \cellcolor{gray!25}OSCR & Acc & H-score & \cellcolor{gray!25}OSCR & Acc & H-score & \cellcolor{gray!25}OSCR \\

\midrule

TCL~\cite{huang2023twin}  & 68.17 & 62.70 & \cellcolor{gray!25}67.56 & 64.48 & 67.84 & \cellcolor{gray!25}61.30 & 45.80 & 35.84 & \cellcolor{gray!25}35.68 & 22.29 & 9.28 & \cellcolor{gray!25}3.54 & 50.19 & 43.92 & \cellcolor{gray!25} 42.02 \\

NPN~\cite{sheng2024adaptive}  & 27.71 & 10.79 & \cellcolor{gray!25}8.48 & 39.34 &  33.88 & \cellcolor{gray!25}32.40 & 40.39 & 44.98 & \cellcolor{gray!25}34.61 & 20.80 & 26.62 & \cellcolor{gray!25}19.84 & 32.06 & 29.07 & \cellcolor{gray!25} 23.83 \\

BadLabel~\cite{zhang2024badlabel}  & 51.62 & 61.25 & \cellcolor{gray!25}50.41 & 32.52 & 37.26 & \cellcolor{gray!25}26.67 & 43.58 & 55.85 & \cellcolor{gray!25}42.45 & 31.60 & 28.24 & \cellcolor{gray!25}26.93 & 39.83 & 45.65 & \cellcolor{gray!25} 36.62\\

DISC~\cite{li2023disc} & 57.35 & 52.08 & \cellcolor{gray!25}44.73 & 38.02 & 37.36 & \cellcolor{gray!25}28.09 & 30.84 & 31.69 & \cellcolor{gray!25}22.68 & 20.72 & 16.32 & \cellcolor{gray!25}16.20 & 36.73 & 34.36 & \cellcolor{gray!25} 27.93 \\

LSL~\cite{kim2024learning}  & 65.99 & 68.68 & \cellcolor{gray!25}62.18 & 55.66 & 49.85 & \cellcolor{gray!25}44.97 & 49.20 & 44.94 & \cellcolor{gray!25}36.86 & 29.90 & 6.91 & \cellcolor{gray!25}19.18 & 50.19 & 42.60 & \cellcolor{gray!25} 40.80 \\
PLM~\cite{zhao2024estimating}  & 57.67 & 50.81 & \cellcolor{gray!25}51.87 & 51.41 & 46.57 & \cellcolor{gray!25}41.82 & 39.66 & 39.47 & \cellcolor{gray!25}31.89 & 19.95 & 12.93 & \cellcolor{gray!25}15.75 & 42.17 & 37.45 & \cellcolor{gray!25} 35.33 \\
\bottomrule
\end{tabular}

}
\caption{Results (\%) of PACS on ViT-Base~\cite{dosovitskiy2021an}.
The open-set ratio is $6{:}1$ and symmetric label noise with ratio $50\%$ is selected.}

\label{tab:pacs_vit_50}
%\vskip-2ex
\end{table*}

\begin{table*}[t!]

\centering
\resizebox{1.\linewidth}{!}{
\begin{tabular}{l|ccc|ccc|ccc|ccc|ccc}
\toprule
& \multicolumn{3}{c|}{\textbf{Photo (P)}} & \multicolumn{3}{c|}{\textbf{Art (A)}} & \multicolumn{3}{c|}{\textbf{Cartoon (C)}} & \multicolumn{3}{c|}{\textbf{Sketch (S)}} & \multicolumn{3}{c}{\textbf{Avg}} \\
\textbf{Method} & Acc & H-score & \cellcolor{gray!25}OSCR & Acc & H-score & \cellcolor{gray!25}OSCR & Acc & H-score & \cellcolor{gray!25}OSCR & Acc & H-score & \cellcolor{gray!25}OSCR & Acc & H-score & \cellcolor{gray!25}OSCR \\

\midrule

TCL~\cite{huang2023twin}  & 22.70 & 25.47 & \cellcolor{gray!25}16.55 & 22.26 & 23.14 & \cellcolor{gray!25}14.42 & 19.75 & 14.23 & \cellcolor{gray!25}6.53 & 28.52 & 10.92 & \cellcolor{gray!25}11.54 & 23.31 & 18.44 & \cellcolor{gray!25} 12.26 \\

NPN~\cite{sheng2024adaptive} & 15.27 & 15.29 & \cellcolor{gray!25}7.9 & 16.07 & 10.88 & \cellcolor{gray!25}11.29 & 19.96 & 17.19 & \cellcolor{gray!25}8.06 & 20.48 & 1.12 & \cellcolor{gray!25}18.43 & 17.95 & 11.12 & \cellcolor{gray!25} 11.42 \\

BadLabel~\cite{zhang2024badlabel}  & 13.33 & 20.96 & \cellcolor{gray!25}13.32 & 24.77 & 32.36 & \cellcolor{gray!25}21.98 & 23.98 & 32.76 & \cellcolor{gray!25}21.91 & 21.60 & 2.36 & \cellcolor{gray!25}\textbf{21.31} & 20.92 & 22.11 & \cellcolor{gray!25} 19.63 \\

DISC~\cite{li2023disc} &22.37 & 1.92 & \cellcolor{gray!25}11.07 & 24.58 & 19.66 & \cellcolor{gray!25}12.81 & 24.08 & 13.40 & \cellcolor{gray!25}14.20 & 20.03 & 4.56 & \cellcolor{gray!25}12.61 & 22.77 & 9.89 & \cellcolor{gray!25} 12.67 \\

LSL~\cite{kim2024learning}  & 10.02 & 3.64 & \cellcolor{gray!25}5.89 & 26.27 & 10.83 & \cellcolor{gray!25}17.29 & 31.87 & 27.63 & \cellcolor{gray!25}19.75 & 25.39 & 8.09 & \cellcolor{gray!25}7.83 & 23.39 & 12.55 & \cellcolor{gray!25} 12.69 \\
PLM~\cite{zhao2024estimating}  & 29.32 & 5.93 & \cellcolor{gray!25}1.34 & 26.77 & 28.58 & \cellcolor{gray!25}18.39 & 25.27 & 25.69 & \cellcolor{gray!25}16.89 & 24.22 & 3.33 & \cellcolor{gray!25}2.25 & 26.40 & 15.88 & \cellcolor{gray!25} 9.72 \\
\bottomrule
\end{tabular}
}
\caption{Results (\%) of PACS on ViT-Base~\cite{dosovitskiy2021an}.
The open-set ratio is $6{:}1$ and symmetric label noise with ratio $80\%$ is selected.}

\label{tab:pacs_vit_80}
%\vskip-2ex
\end{table*}

\begin{table*}[t!]

\centering
\resizebox{1.\linewidth}{!}{
\begin{tabular}{l|ccc|ccc|ccc|ccc|ccc}
\toprule
& \multicolumn{3}{c|}{\textbf{Photo (P)}} & \multicolumn{3}{c|}{\textbf{Art (A)}} & \multicolumn{3}{c|}{\textbf{Cartoon (C)}} & \multicolumn{3}{c|}{\textbf{Sketch (S)}} & \multicolumn{3}{c}{\textbf{Avg}} \\
\textbf{Method} & Acc & H-score & \cellcolor{gray!25}OSCR & Acc & H-score & \cellcolor{gray!25}OSCR & Acc & H-score & \cellcolor{gray!25}OSCR & Acc & H-score & \cellcolor{gray!25}OSCR & Acc & H-score & \cellcolor{gray!25}OSCR \\

\midrule

TCL~\cite{huang2023twin}  & 51.21 & 56.51 & \cellcolor{gray!25}48.47 & 46.65 & 51.35 & \cellcolor{gray!25}40.80 & 31.92 & 30.94 & \cellcolor{gray!25}22.61 & 25.87 & 20.94 & \cellcolor{gray!25}12.04 & 38.91 & 39.94 & \cellcolor{gray!25}30.98 \\

NPN~\cite{sheng2024adaptive}  & 27.63 & 8.86 & \cellcolor{gray!25}6.32 & 32.40 & 19.00 & \cellcolor{gray!25}15.72 & 21.66 & 16.87 & \cellcolor{gray!25}14.25 & 20.48 & 26.15 & \cellcolor{gray!25}16.50 & 25.54 & 17.72 & \cellcolor{gray!25}13.20 \\

BadLabel~\cite{zhang2024badlabel}  & 46.12 & 58.96 & \cellcolor{gray!25}45.77 & 35.33 & 34.30 & \cellcolor{gray!25}34.83 & 26.92 & 39.52 & \cellcolor{gray!25}26.18 & 20.70 & \textbf{31.32} & \cellcolor{gray!25}20.66 & 32.27 & 41.03 & \cellcolor{gray!25}31.86 \\

DISC~\cite{li2023disc} & 47.17 & 17.08 & \cellcolor{gray!25}9.77 & 24.64 & 19.96 & \cellcolor{gray!25}10.75 & 20.22 & 12.84 & \cellcolor{gray!25}9.22 & 23.93 & 4.31 & \cellcolor{gray!25}15.10 & 28.99 & 13.55 & \cellcolor{gray!25}11.21 \\

LSL~\cite{kim2024learning}  & 49.52 & 5.63 & \cellcolor{gray!25}15.99 & 37.52 & 34.68 & \cellcolor{gray!25}26.68 & 28.16 & 27.60 & \cellcolor{gray!25}18.13 & 28.16 & 27.60 & \cellcolor{gray!25}18.13 & 35.84 & 23.88 & \cellcolor{gray!25}19.73 \\
PLM~\cite{zhao2024estimating}  & 20.51 & 23.01 & \cellcolor{gray!25}18.15 & 20.17 & 13.41 & \cellcolor{gray!25}8.08 & 37.85 & 33.88 & \cellcolor{gray!25}26.43 & 20.59 & 24.16 & \cellcolor{gray!25}17.26 & 24.78 & 23.62 & \cellcolor{gray!25}17.48 \\
\bottomrule
\end{tabular}

}
\caption{Results (\%) of PACS on ViT-Base~\cite{dosovitskiy2021an}. The open-set ratio is $6{:}1$ and asymmetric label noise with ratio $50\%$ is selected.}

\label{tab:pacs_vit_50a}
%\vskip-2ex
\end{table*}
\begin{table*}[t!]

\centering
\resizebox{1.\linewidth}{!}{
\begin{tabular}{l|ccc|ccc|ccc|ccc|ccc}
\toprule
& \multicolumn{3}{c|}{\textbf{mnist}} & \multicolumn{3}{c|}{\textbf{mnist$_m$}} & \multicolumn{3}{c|}{\textbf{syn}} & \multicolumn{3}{c|}{\textbf{svhn}} & \multicolumn{3}{c}{\textbf{Avg}} \\
\textbf{Method} & Acc & H-score & \cellcolor{gray!25}OSCR & Acc & H-score & \cellcolor{gray!25}OSCR & Acc & H-score & \cellcolor{gray!25}OSCR & Acc & H-score & \cellcolor{gray!25}OSCR & Acc & H-score & \cellcolor{gray!25}OSCR \\

\midrule
NPN~\cite{sheng2024adaptive} & 82.28 & 29.42 & \cellcolor{gray!25}70.68 & 32.56 & 28.22 & \cellcolor{gray!25}22.74 & 21.78 & 22.54 & \cellcolor{gray!25}14.41 & 17.33 & 11.95 & \cellcolor{gray!25}9.89 & 38.49 & 23.03 & \cellcolor{gray!25}29.43 \\

BadLabel~\cite{zhang2024badlabel}  & 63.25 & 52.49 & \cellcolor{gray!25}61.18 & 30.00 & 34.82 & \cellcolor{gray!25}27.42 & 50.17 & 51.19 & \cellcolor{gray!25}46.06 & 41.82 & 16.75 & \cellcolor{gray!25}41.82 & 46.31 & 38.81 & \cellcolor{gray!25}44.12 \\

\midrule 

ODGNet~\cite{bose2023beyond}& 90.33 & 50.84 & \cellcolor{gray!25}71.38 & 59.28 & 26.10 & \cellcolor{gray!25}43.83 & 70.11 & 53.83 & \cellcolor{gray!25}49.54 & 55.81 & 32.71 & \cellcolor{gray!25}39.54 & 68.88 & 40.87 & \cellcolor{gray!25}51.07 \\

MLDG \cite{chen2021adversarial} & 90.67 & 27.60 & \cellcolor{gray!25}80.46 & 57.89 & 48.38 & \cellcolor{gray!25}42.77 & 60.33 & 39.95 & \cellcolor{gray!25}41.30 & 53.75 & 6.08 & \cellcolor{gray!25}35.20 & 65.66 & 30.50 & \cellcolor{gray!25}49.93 \\

MEDIC-cls~\cite{wang2023generalizable} & 22.08 & 8.71 & \cellcolor{gray!25}5.89 & 21.33 & 20.09 & \cellcolor{gray!25}9.94 & 23.24 & 9.83 & \cellcolor{gray!25}13.28 & 13.72 & 10.29 & \cellcolor{gray!25}6.63 & 20.09 & 12.23 & \cellcolor{gray!25}8.94 \\

MEDIC-bcls~\cite{wang2023generalizable} & 22.08 & 12.31 & \cellcolor{gray!25}5.57 & 21.33 & 16.19 & \cellcolor{gray!25}10.11 & 23.24 & 12.64 & \cellcolor{gray!25}5.41 & 13.72 & 13.47 & \cellcolor{gray!25}5.94 & 20.09 &  13.65 & \cellcolor{gray!25}6.76 \\
\bottomrule
\end{tabular}}
\caption{Results ($\%$) of DigitsDG on ConvNet~\cite{zhou2021domain}, where symmetric label noise with ratio $20\%$ is selected.}
\label{tab:dg_20_append}
%\vskip-2ex
\end{table*}

\begin{table*}[t!]

\centering
\resizebox{1.\linewidth}{!}{
\begin{tabular}{l|ccc|ccc|ccc|ccc|ccc}
\toprule
& \multicolumn{3}{c|}{\textbf{mnist}} & \multicolumn{3}{c|}{\textbf{mnist$_m$}} & \multicolumn{3}{c|}{\textbf{syn}} & \multicolumn{3}{c|}{\textbf{svhn}} & \multicolumn{3}{c}{\textbf{Avg}} \\
\textbf{Method} & Acc & H-score & \cellcolor{gray!25}OSCR & Acc & H-score & \cellcolor{gray!25}OSCR & Acc & H-score & \cellcolor{gray!25}OSCR & Acc & H-score & \cellcolor{gray!25}OSCR & Acc & H-score & \cellcolor{gray!25}OSCR \\

\midrule	

NPN~\cite{sheng2024adaptive} & 68.11 & 25.80 & \cellcolor{gray!25}49.14 & 28.31 & 28.29 & \cellcolor{gray!25}18.65 & 45.78 & 32.17 & \cellcolor{gray!25}31.15 & 28.42 & 1.38 & \cellcolor{gray!25}16.37 & 42.66 & 21.91 & \cellcolor{gray!25}28.83 \\

BadLabel~\cite{zhang2024badlabel}  & 63.31 &35.12  & \cellcolor{gray!25}57.47& 42.28 & 42.69 & \cellcolor{gray!25}38.44 & 21.36 & 25.07 & \cellcolor{gray!25}18.30 & 19.83 & 4.88 & \cellcolor{gray!25}19.28 & 36.70 & 26.94 & \cellcolor{gray!25}33.37 \\

\midrule

ODGNet~\cite{bose2023beyond}& 71.25 & 22.26 & \cellcolor{gray!25}49.81 & 59.22 & 36.88 & \cellcolor{gray!25}44.86 & \textbf{61.39} & \textbf{43.65} & \cellcolor{gray!25}\textbf{42.41} & \textbf{51.75} & 19.70 & \cellcolor{gray!25}\textbf{35.38} & 60.90 & 30.62 & \cellcolor{gray!25}43.12 \\

MLDG~\cite{shu2019meta} & 62.72 & 54.21 & \cellcolor{gray!25}50.52 & 48.94 & 41.54 & \cellcolor{gray!25}35.35 & 43.53  & 33.15 & \cellcolor{gray!25}29.09 & 24.61 & 6.48 & \cellcolor{gray!25}14.29 & 44.95 & 33.85 & \cellcolor{gray!25}32.31 \\

MEDIC-cls~\cite{wang2023generalizable} & 22.39 & 0.17 & \cellcolor{gray!25}2.81 & 25.89 & 22.88 & \cellcolor{gray!25}12.16 & 12.78 & 12.10 & \cellcolor{gray!25}4.73 & 7.61 & 5.70 & \cellcolor{gray!25}2.35 & 17.17 & 10.21 & \cellcolor{gray!25}5.51 \\
\bottomrule
\end{tabular}

}
\caption{Results (\%) of DigitsDG on ConvNet~\cite{zhou2021domain}, where symmetric label noise with ratio $50\%$ is selected.}

\label{tab:dg_50}
%\vskip-2ex
\end{table*}

\begin{table*}[t!]

\centering
\resizebox{1.\linewidth}{!}{
\begin{tabular}{l|ccc|ccc|ccc|ccc|ccc}
\toprule
& \multicolumn{3}{c|}{\textbf{mnist}} & \multicolumn{3}{c|}{\textbf{mnist$_m$}} & \multicolumn{3}{c|}{\textbf{syn}} & \multicolumn{3}{c|}{\textbf{svhn}} & \multicolumn{3}{c}{\textbf{Avg}} \\
\textbf{Method} & Acc & H-score & \cellcolor{gray!25}OSCR & Acc & H-score & \cellcolor{gray!25}OSCR & Acc & H-score & \cellcolor{gray!25}OSCR & Acc & H-score & \cellcolor{gray!25}OSCR & Acc & H-score & \cellcolor{gray!25}OSCR \\

\midrule

NPN~\cite{sheng2024adaptive} & 16.67 & 0.01 & \cellcolor{gray!25}9.65 & 18.61 & 13.38 & \cellcolor{gray!25}10.15 & 17.78 & 18.12 & \cellcolor{gray!25}9.29 & 16.67 & 0.55 & \cellcolor{gray!25}8.21 & 17.43 & 8.02 & \cellcolor{gray!25}9.33 \\

BadLabel~\cite{zhang2024badlabel} & 18.58 & 5.84 & \cellcolor{gray!25}7.39 &17.39  & 15.13 & \cellcolor{gray!25}8.99 & 16.67 & 0.41 & \cellcolor{gray!25}8.36 & 17.11 & 0.22 & \cellcolor{gray!25}8.44 & 17.44 & 5.40 & \cellcolor{gray!25}8.30 \\

\midrule

ODGNet~\cite{bose2023beyond} & 16.19 & 1.51  & \cellcolor{gray!25}10.37 & 17.28 & 11.80 & \cellcolor{gray!25}10.64 & 18.47 & 17.04 & \cellcolor{gray!25}8.49 & 16.72 & 14.89 & \cellcolor{gray!25}8.59 &17.17 & 11.31 & \cellcolor{gray!25} 9.52\\

MLDG~\cite{shu2019meta} & 16.06 & 6.70 & \cellcolor{gray!25}9.69 & 18.58 & 3.26 & \cellcolor{gray!25}9.27 & 16.94 & 6.81 & \cellcolor{gray!25}8.20 & 17.33 & 6.88 & \cellcolor{gray!25}8.84 & 17.23 & 5.91 & \cellcolor{gray!25} 9.00\\

MEDIC-cls~\cite{wang2023generalizable} & 21.17 & 3.51 & \cellcolor{gray!25}11.37 & 18.75 & 16.52 & \cellcolor{gray!25}7.77 & 15.81 & 4.48 & \cellcolor{gray!25}7.54 & 17.11  & 4.86 & \cellcolor{gray!25}8.64 & 18.21 & 7.34 & \cellcolor{gray!25}8.83 \\

MEDIC-bcls~\cite{wang2023generalizable} & 21.17 & 7.21 & \cellcolor{gray!25}8.32 & 16.83 & 13.25 & \cellcolor{gray!25}8.81 & 15.81 & 4.44 & \cellcolor{gray!25}7.70 & 17.11 & 4.96 & \cellcolor{gray!25}8.38 & 17.73 & 7.47 & \cellcolor{gray!25}8.30 \\
\bottomrule
\end{tabular}

}
\caption{Results (\%) of DigitsDG on ConvNet~\cite{zhou2021domain}, where symmetric label noise with ratio $80\%$ is selected.}
\label{tab:dg_80}
%\vskip-2ex
\end{table*}

\begin{table*}[h!]

\centering
\resizebox{1.\linewidth}{!}{
\begin{tabular}{l|ccc|ccc|ccc|ccc|ccc}
\toprule
& \multicolumn{3}{c|}{\textbf{mnist}} & \multicolumn{3}{c|}{\textbf{mnist$_m$}} & \multicolumn{3}{c|}{\textbf{syn}} & \multicolumn{3}{c|}{\textbf{svhn}} & \multicolumn{3}{c}{\textbf{Avg}} \\
\textbf{Method} & Acc & H-score & \cellcolor{gray!25}OSCR & Acc & H-score & \cellcolor{gray!25}OSCR & Acc & H-score & \cellcolor{gray!25}OSCR & Acc & H-score & \cellcolor{gray!25}OSCR & Acc & H-score & \cellcolor{gray!25}OSCR \\

\midrule

NPN~\cite{sheng2024adaptive} & 71.08 & 24.46 & \cellcolor{gray!25}61.45 & 54.58 & 41.21 & \cellcolor{gray!25}38.79 & 55.92 & 42.95 & \cellcolor{gray!25}35.03 & 51.97 & 11.28 & \cellcolor{gray!25}34.23 &58.39 & 29.98 & \cellcolor{gray!25}42.38 \\

BadLabel~\cite{zhang2024badlabel}  & 53.00 & 38.77 & \cellcolor{gray!25}43.34 & 33.94 & 27.79 & \cellcolor{gray!25}30.06 & 37.64 & 44.59 & \cellcolor{gray!25}33.47 & 27.61 & 20.90 & \cellcolor{gray!25}26.18 & 38.05 & 33.01 & \cellcolor{gray!25}33.26 \\

\midrule

ODGNet~\cite{bose2023beyond} & 53.67 & 48.23 & \cellcolor{gray!25}41.31 & 39.36 & 27.19 & \cellcolor{gray!25}28.08 & 51.17 & 24.92 & \cellcolor{gray!25}34.91 & 39.64 & 5.51 & \cellcolor{gray!25}25.32 &45.96&	26.46	  & \cellcolor{gray!25}32.41   \\

MLDG~\cite{shu2019meta} & 68.17 & 23.34 & \cellcolor{gray!25}55.68 & 56.47 & 40.89 & \cellcolor{gray!25}40.97 & 56.31 & 41.94 & \cellcolor{gray!25}38.02 & 47.81 & 11.35 & \cellcolor{gray!25}31.60 &57.19 &	29.38	  & \cellcolor{gray!25}41.57  \\

MEDIC-cls~\cite{wang2023generalizable} & 19.86 & 14.41 & \cellcolor{gray!25}11.22 & 19.75 & 15.93 & \cellcolor{gray!25}8.03 & 9.83 & 8.28 & \cellcolor{gray!25}2.88 & 16.31 & 5.78 & \cellcolor{gray!25}7.36 &16.44  &	11.10	  & \cellcolor{gray!25}7.37 \\

MEDIC-bcls~\cite{wang2023generalizable} & 19.86 & 21.31 & \cellcolor{gray!25}11.22 & 19.75 & 14.66 & \cellcolor{gray!25}6.73 & 9.83 & 8.82 & \cellcolor{gray!25}2.58 & 16.31 & 13.40 & \cellcolor{gray!25}7.39 &16.44  &	14.55	  & \cellcolor{gray!25}6.98 \\
\bottomrule
\end{tabular}
}
\caption{Results (\%) of DigitsDG on ConvNet~\cite{zhou2021domain}, where asymmetric label noise with ratio $50\%$ is selected.}

\label{tab:dg_50a}
%\vskip-2ex
\end{table*}

\section{Ablation of Unsupervised Clustering for Clean-Noisy Partition}
In Tab.~\ref{tab:unsup}, we present ablation experiments on the unsupervised clustering approach for label-clean/noisy set partitioning. Separation correctness is evaluated using accuracy, where a binary indicator serves as ground truth, denoting whether the current label matches the original unperturbed label. We compare our method with two variants, \ie., GMM and FINCH, where we directly apply GMM and FINCH on the recorded loss to achieve binary clustering. From the experimental results, we can observe that our approach generally outperforms those two variants. FINCH~\cite{finch} shows comparable performance with our approach on the symmetric label noise ratio of $20\%$, while our approach outperforms FINCH by large margins on the other label noise settings, demonstrating that the combination of the FINCH and GMM classifier is more robust to severe label noise. We further deliver more analysis for the sensitivity of the proposed HyProMeta to the clean/noisy partition.  On PACS with art painting as the target domain and $50\%$ label noise, reducing clustering accuracy from $92.25\%$ to $42.76\%$ and leading to a smaller OSCR drop from $59.58\%$ to $46.97\%$. While performance is affected, the method remains robust due to selective clean sample usage, evidential pseudo-labeling, and meta-learning regularization.

\begin{table*}[t!]

\centering
\resizebox{\linewidth}{!}{
\begin{tabular}{l|ccc|ccc|ccc|ccc}
\toprule
& \multicolumn{3}{c|}{\textbf{20\% sym}} & \multicolumn{3}{c|}{\textbf{50\% sym}} & \multicolumn{3}{c|}{\textbf{80\% sym}} & \multicolumn{3}{c}{\textbf{50\% asym}} \\
\textbf{Method} & Acc & H-score & \cellcolor{gray!25}OSCR & Acc & H-score & \cellcolor{gray!25}OSCR & Acc & H-score & \cellcolor{gray!25}OSCR & Acc & H-score & \cellcolor{gray!25}OSCR \\

\midrule
HyProMeta~\cite{peng2024mitigating} & 56.61 & 37.75 & \cellcolor{gray!25}28.86 & 47.90 & 18.56 & \cellcolor{gray!25} 30.64& 34.72& 20.82 & \cellcolor{gray!25}24.14 & 36.77 & 15.19 & \cellcolor{gray!25}24.71\\
\midrule
Ours & \textbf{58.47} & \textbf{37.86} & \cellcolor{gray!25}\textbf{30.25} & \textbf{50.10} & \textbf{33.99} & \cellcolor{gray!25} \textbf{33.50}&\textbf{49.55} & \textbf{22.89} & \cellcolor{gray!25}\textbf{32.40} & \textbf{40.73} & \textbf{37.49} & \cellcolor{gray!25}\textbf{28.50}\\  
\bottomrule
\end{tabular}}

\caption{Experimental results on TerriaINC dataset from DomainBed.}
\label{tab:terrainc}

\end{table*}

\section{Training overhead and computation cost of DC-CRFM}

The number of parameters of our method is $\sim 215.6M$ during training, where DC-CRFM takes $\sim 129.6M$ due to its encoder-decoder structure for the generation of residuals, and $\sim 86.0M$ during testing when we use ViT-Base~\cite{dosovitskiy2021an} as backbone, since DC-CRFM only participates in training. The whole training procedure takes $\sim 5h$ on PACS when we use one A100 GPU and ViT-Base as backbone.

\begin{table}[t!]
\centering
\resizebox{0.6\linewidth}{!}{\begin{tabular}{l|cccc}
\toprule
\textbf{Method} & \multicolumn{1}{l}{\textbf{20\% sym}} & \multicolumn{1}{l}{\textbf{50\% sym}} & \multicolumn{1}{l}{\textbf{80\% sym}} & \textbf{50\% asym} \\
\midrule
GMM                        & 72.11                        & 87.83                        & 54.56                        & 50.49        \\
FINCH                      & 90.39                        & 85.29                        & 50.02                        & 38.71        \\
\midrule
Ours                       & 90.04                        & 92.25    & 56.05                        & 52.91      \\
\bottomrule

\end{tabular}}
\caption{Ablation experiments on PACS \textit{art painting} using ResNet18~\cite{he2016deep} as backbone for the unsupervised clustering approach regarding the label-clean/noisy sets partition. The performance is evaluated by the accuracy computed over the partitioned sample set using a binary indicator of whether the uncleaned label matches the original label for each sample.}
\label{tab:unsup}
\end{table}

\begin{table}[t!]

\centering
\resizebox{0.6\linewidth}{!}{
\begin{tabular}{l|c|ccc|ccc}
\toprule
&&\multicolumn{3}{c|}{\textbf{PACS (Photo)}} & \multicolumn{3}{c}{\textbf{DigitsDG (mnist)}} \\
\textbf{Method} &\textbf{\#Params}  & Acc & H-score & \cellcolor{gray!25}OSCR & Acc & H-score & \cellcolor{gray!25}OSCR \\
\midrule
DiT-S & 30.98M& 77.71& 75.29& \cellcolor{gray!25}70.07 & 74.86 &4.45 & \cellcolor{gray!25}63.42 \\
DiT-B & 129.60M&82.39& 81.52& \cellcolor{gray!25}78.68 & 85.97 & 64.79& \cellcolor{gray!25}69.88 \\ 
DiT-L &435.90M & 77.46 & 65.01 & \cellcolor{gray!25} 63.65 & 78.83 & 22.96& \cellcolor{gray!25} 66.43\\ 
\bottomrule
\end{tabular}}
\caption{Ablation regarding the scalability of DC-CRFM using different sizes of DiT. Experiments are conducted on PACS dataset (test domain: \textit{Photo}) and DigitsDG dataset (test domain: \textit{MNIST}).}
\label{tab:scability}
\end{table}

\section{Further Clarification regarding the Generalizability to Other Datasets}
We conduct further experiments on TerraInc dataset~\cite{beery2018recognition} from DomainBed~\cite{peng2019moment} with open-set ratio ($8{:}2$). The results are reported in Tab.~\ref{tab:terrainc}, where we find our approach still outperforms the current best approach, HyProMeta.
Across all noise conditions, the proposed method (Ours) outperforms HyProMeta in all metrics. Notably, under $50\%$ symmetric noise, it achieves a significant H-score gain of $15.43\%$ and OSCR gain of $2.86\%$, indicating improved robustness in separating clean/noisy samples and generalizing to unseen categories. Even under high-noise settings ($80\%$ symmetric and $50\%$ asymmetric), our method maintains superior OSCR and H-score, validating its effectiveness in tackling the OSDG-NL task.

\section{Scaling of DiT}
We provide the scalability evaluation in Tab.~\ref{tab:scability}, where we find that DiT-B~\cite{Peebles2022DiT} works the best compared to DIT-S/L across different datasets, and we also adopt DiT-B in our experiments. For PACS on the test domain \textit{Photo}, DiT-B achieves the best results (Acc: $82.39\%$, H-score: $81.52\%$, OSCR: $78.68\%$), showing that scaling from DiT-S to DiT-B improves performance. However, further increasing the model size to DiT-L results in performance degradation, especially in H-score and OSCR.

For DigitsDG on the test domain \textit{MNIST}, the gap is even more pronounced. DiT-B again performs the best (Acc: $85.97\%$, H-score: $64.79\%$, OSCR: $69.88\%$), whereas DiT-L suffers a sharp drop in H-score ($22.96\%$) despite having the highest parameter count. This indicates that DiT-B offers the best balance between model complexity and generalization for DC-CRFM.

\end{document}